\pdfoutput=1
\documentclass[11pt]{article}

\usepackage{EMNLP2022}

\usepackage{times}
\usepackage{latexsym}

\usepackage[T1]{fontenc}
\usepackage[utf8]{inputenc}
\usepackage{xcolor,colortbl}
\usepackage{microtype}
\usepackage{graphicx}

\usepackage{inconsolata}
\usepackage{booktabs, subcaption}

\title{An Automatic Evaluation\\ of the WMT22 General Machine Translation Task}

\author{Benjamin Marie \\
  \texttt{nlp@benjaminmarie.com}}

\begin{document}
\maketitle
\begin{abstract}
This report presents an automatic evaluation of the general machine translation task of the Seventh Conference on Machine Translation (WMT22). It evaluates a total of 185 systems for 21 translation directions including high-resource to low-resource language pairs and from closely related to distant languages. This large-scale automatic evaluation highlights some of the current limits of state-of-the-art machine translation systems. It also shows how automatic metrics, namely chrF, BLEU, and COMET, can complement themselves to mitigate their own limits in terms of interpretability and accuracy.
\end{abstract}

\section{Introduction}
In its 2022 edition, the Conference on Machine Translation (WMT22)\footnote{\url{https://www.statmt.org/wmt22/}} organized a new ``general'' machine translation shared task. The participants in this task were asked to prepare machine translation (MT) systems for translating texts from potentially any domains and most likely ``news, social, conversational, and ecommerce.''

For this task, WMT22 proposed a large variety of language pairs: Czech$\leftrightarrow$English (cs-en), Czech$\leftrightarrow$Ukrainian (cs-uk), German$\leftrightarrow$English (de-en), French$\leftrightarrow$German (fr-de), English$\rightarrow$Croatian (en-hr), English$\leftrightarrow$Japanese (en-ja), English$\leftrightarrow$Livonian (en-liv), English$\leftrightarrow$Russian (en-ru), Russian$\leftrightarrow$Yakut (ru-sah), English$\leftrightarrow$Ukrainian (en-uk), and English$\leftrightarrow$Chinese (en-zh). The organizers classified all these language pairs as in Table \ref{tab:langclass} according to the relatedness of the languages and the quantity of resources available for training an MT system. In total, there are 11 language pairs and 21 translation directions.

In this paper, we present an automatic evaluation\footnote{This is not an official paper of WMT22. WMT22 organizers will perform a human evaluation.} of the 185 systems submitted\footnote{It includes the ``Online-*'' systems added by the organizers.} to the general machine translation task. We used the translations officially released by WMT22.\footnote{\url{https://github.com/wmt-conference/wmt22-news-systems}} We analyze the results with different metrics and methods. Our main observations are as follows:

\begin{itemize}
    \item Low-resource distant language pairs remain extremely challenging. The best systems for en$\leftrightarrow$liv and ru$\leftrightarrow$sah are very far from reaching the reference. ja$\leftrightarrow$en and en$\rightarrow$uk also remain challenging despite being medium-resource. Among the high-resource language pairs, the systems built for en$\rightarrow$ru struggled the most to approximate the reference translation.
    \item A BLEU score difference higher than 0.9 is always statistically significant in this task.
    \item A naive combination of all the evaluated systems using minimum Bayes risk (MBR) decoding with COMET can improve the BLEU, chrF and COMET scores.
    \item Only a few systems seem to have exploited document-level features or a context larger than one segment.
    \item Absolute COMET scores are meaningless and must be completed with another metric score to get an estimate of the system's translation quality.
\end{itemize}

\begin{table}
    \centering
    \begin{tabular}{lccc}
    \toprule
         & High-resource & Medium-resource & Low-resource  \\
    \midrule
        Closely related& & cs-uk\\
        Same family & cs-en, de-en, en-ru & de-fr, en-uk & en-hr\\
        Distant & en-zh & en-ja & en-liv, ru-sah  \\
    \bottomrule
    \end{tabular}
    \caption{\label{tab:langclass}Language pairs in terms of resource available for training and language similarity as defined by WMT22.}
    \label{tab:my_label}
\end{table}

The paper is organized as follows. In each section, we present an overview of the main findings with the numerous associated tables given in the appendix. In Section \ref{sec:ranking}, we score and rank the systems with three different metrics. These rankings are completed by a statistical significance testing on all possible pairs of systems in Section \ref{sec:sigtest}. Section \ref{sec:mbr} presents the results of a combination of all the submissions using MBR and an oracle guided by COMET. Section \ref{sec:matchs} presents various statistics of the submitted systems to better highlight how they stand out. For a more in-depth evaluation with COMET, we evaluate all the systems with 5 different COMET models in Section \ref{sec:comet}. Finally, Section \ref{sec:ccl} concludes this evaluation.

\section{Ranking with Metrics}
\label{sec:ranking}
For this evaluation, we will rely on three different automatic metrics:
\begin{itemize}
\item chrF \citep{popovic-2015-chrf}: A tokenization independent metric operating at character-level with a higher correlation with human judgments than BLEU.
\item BLEU \citep{papineni-etal-2002-bleu}: The standard BLEU. 
\item COMET \citep{rei-etal-2020-comet}: A state-of-the-art metric based on a pre-trained language model. We used the default model ``wmt20-comet-da.''
\end{itemize}

Note that chrF and BLEU are merely used for diagnostic purposes and to answer the question: How far are we from reaching particular reference translations? On the other hand, we used COMET mainly to produce more accurate rankings that would better correlate with the human evaluation.

chrF and BLEU scores are computed with SacreBLEU \citep{post-2018-call}.\footnote{\url{https://github.com/mjpost/sacrebleu}} If there are two reference translations available, both are used to compute the chrF and BLEU scores,\footnote{We did not use the reference ``stud'' created by students.} while we only used the first available reference to compute COMET scores. The systems are ranked given their scores but a rank is assigned only to the systems that have been declared ``constrained'' by their authors, i.e., systems that only used the data provided by the organizers. The rankings and scores are given in Section \ref{sec:rankingtab}, from Table \ref{tab:firstranking} to \ref{tab:lastranking}. In these tables, systems with a rank ``n/a'' are systems that are not constrained.

Having two references, we obtain absolute BLEU scores rarely seen in the machine translation research literature with, for instance, 60.9 BLEU points for JDExploreAcademy for cs$\rightarrow$en. Even higher BLEU scores are observed for en$\rightarrow$zh due to the use of smaller tokens that makes the 4-gram matching a much easier task. Such high scores may be misleading about the translation quality. Absolute BLEU scores do not inform us on the translation quality itself, but they do inform us that these systems produce many 4-grams that are in the reference translations.

While chrF and BLEU directly indicate how well the translation matches the references with a score between 0 and 100 points, COMET scores are not bounded between 0 and 100 making the interpretation of absolute scores extremely challenging. For instance, at the extremes, AMU obtains 104.9 COMET points for uk$\rightarrow$cs and AIST obtains -152.7 COMET points for liv$\rightarrow$en. BLEU and chrF absolute scores can be used for diagnostic purposes and answer basic questions: How close are we from the reference with a given tokenization? Is the system likely generating text in the target languages? etc. COMET cannot, but is much more reliable for ranking systems as demonstrated in previous work \citep{rei-etal-2020-comet}.

BLEU and chrF yields the same rankings for only 5 translation directions: uk$\rightarrow$cs, de$\rightarrow$fr, sah$\rightarrow$ru, and ru$\rightarrow$sah. Nonetheless, BLEU and chrF yields ranking that are very similar and find the same best systems for 20 among the 21 translation directions. On the other hand, for 11 among the 21 language pairs, COMET finds a best system that is not among the best systems found by BLEU and chrF. For some system pairs, the difference between BLEU and COMET can be very significant (more on this in Section \ref{sec:sigtest}). For instance for cs$\rightarrow$uk, in terms of score differences, the difference between Online-B and AMU is of 3.6 BLEU points and -5.1 COMET points.

Surprisingly, for some translation directions, constrained systems outperform systems that are not constrained. According to COMET, this is the case for cs$\rightarrow$uk, uk$\rightarrow$cs, de$\rightarrow$en, ja$\rightarrow$en, and en$\rightarrow$ja. 

For some other directions, online systems seem to be better by a large margin. For instance, for de$\rightarrow$fr, Online-W is better than the best constrained system by 18.3 BLEU points.

\section{Statistical Significance Testing}
\label{sec:sigtest}
We would like to measure how reliable is the conclusion that a system is better than another one according to some metric. In other words, we would like to test whether the difference between systems' metric scores is statistically significant. There are a several tools and techniques to perform statistical significance testing. For this evaluation, we chose the most commonly used: paired bootstrap resampling \citep{koehn-2004-statistical}. We used the implementation in SacreBLEU with its default parameters for chrF and BLEU. For COMET, we used the default parameters of ``comet-compare.'' Results are reported in Section \ref{sec:sigtab}, from Table \ref{tab:firstsig} to \ref{tab:lastsig}. In these tables, the difference between the metric scores is reported for each system pair. The background color is darker for more significant differences (lower p-value) and the score difference is underline if the p-value is below 0.05.

A first interesting observation we have is that a difference in BLEU higher than 0.9 points (cs$\rightarrow$uk) is always significant (p-value $<0.05$). In chrF, the largest difference that is not significant is 0.6 points (en$\rightarrow$zh), while it reaches 2.6 points (liv$\rightarrow$en) for COMET.\footnote{Note that this would highly vary depending on the model used with COMET.}

The significance of the difference between systems can be very different depending on the metric. Overall, there is no clear pattern. For some translation directions, one metric will find a cluster of similar systems while another metric will find that they are all significantly different. This is well-illustrated with en$\rightarrow$ru where BLEU finds a cluster of similar systems while COMET finds that they are almost all significantly different.

The three metrics only agree on a system that is significantly better than all the others for a few translation directions: cs$\rightarrow$en (Online-W), fr$\rightarrow$de (Online-W), en$\rightarrow$liv (TAL-SJTU), sah$\rightarrow$ru (Online-G), and en$\rightarrow$uk (Online-B).

It will be interesting to observe how well the clusters of systems found with statistical significance testing match the clusters found with WMT22 human evaluation.

\section{Naive MBR and Oracle Combinations}
\label{sec:mbr}
As shown by the tables presented in Section \ref{sec:sigtest}, most of the evaluated translations are significantly different according to COMET. One way to exploit this diversity is to combine the translations to generate a better translation. There are many methods to combine machine translation outputs. In this section, we only present two different methods for combination:

\begin{itemize}
    \item Minimum Bayes Risk (MBR): Naively use all the evaluated translations and performs MBR decoding using COMET.
    \item Oracle: Select the best segments among all the evaluated translations according to their COMET score. This is an oracle experiment since we use the COMET scores computed with a reference translation. This experiment is mainly to get an idea of what would be the best translation obtained from the combination of all the systems.
\end{itemize}

The results of these combinations are reported in Section \ref{sec:mbrtab}, from Table \ref{tab:firstmbr} to \ref{tab:lastmbr}. These MBR and oracle combinations rely on COMET but we also report on the chrF and BLEU scores of the resulting translations. ``Baseline'' denotes the score of the best single system for each metric.

As we observe in the tables, the MBR combination improves on all the three metrics for cs$\rightarrow$uk, uk$\rightarrow$cs (with 1.9 BLEU points of improvement), de$\rightarrow$en, en$\rightarrow$de, and liv$\rightarrow$en.

However, for some translation directions, the MBR combination largely fails in improving the metric scores and may even decrease the COMET scores compared to the best single system. For instance, the MBR combination leads to lower COMET scores for cs$\rightarrow$en, de$\rightarrow$fr, ru$\rightarrow$sah, sah$\rightarrow$ru, and uk$\rightarrow$en. This is presumably due to the presence of several translations of a very different and lower quality compared to the best single translation.

On the other hand, the oracle combination improves the scores on all metrics for all translation directions. This shows that BLEU and chrF tend to agree, mostly, that a segment with a higher COMET score is better.

%Cannot see a world where MT researchers report on negative scores....

\section{Sentence Matching and Normalization}
\label{sec:matchs}
In this section, we look at various statistics to try to identify some salient characteristics of the evaluated systems. Statistics are presented in Section \ref{sec:matchtab} from Table \ref{tab:firstmatch} to \ref{tab:lastmatch}.

\paragraph{Exact Match} We counted how many segments in the system translation are identical to their corresponding segments in the reference translation.

\paragraph{Self Mismatch} We counted the number of non-unique source segments translated differently by the system and the reference translation. It leads to interesting observations. Most systems always identically translate the same source segments, but there are some notable exceptions. For instance, Online-W generated different translations for the same segment for almost all its translation directions. For 47 of the 122 non-unique source segments of the en$\rightarrow$ja, Online-W generates a different translation. Among the other submissions to this translation direction, only LanguageX (1) and NAIST-NICT-TIT (22) generated different translations for the same source segments. In contrast, the human translator(s?) produced 63 different translations for the 122 non-unique segments. This is the highest number among all the translation directions. The case of Japanese is particularly interesting here because in this language the context can frequently and radically lead to different translations for the same text. Getting different translations for the same segment is a clue that the system either takes into account document-level information and/or use sampling strategies with some randomness at inference time. CUNI-DocTransformer is another interesting example for cs$\rightarrow$en since it translates differently as many non-unique source segments (7) as the reference translation, probably by exploiting document-level features as suggested by the name of the system. For 10 translation directions, the reference translation translates all the non-unique segments identically.

\paragraph{Sentence Normalization} We also normalized the translation outputs to observe how BLEU and COMET are sensitive to changes in the punctuation marks and encoding issues. It can also highlight whether a system relied on some special post-processing to increase the metric scores. For normalization, we used the following sequence of Moses \citep{koehn-etal-2007-moses} scripts:\footnote{\url{https://github.com/moses-smt/mosesdecoder/tree/master/scripts}} \texttt{tokenizer/replace-unicode-punctuation.perl | tokenizer/normalize-punctuation.perl -l <target\_language> | /tokenizer/remove-non-printing-char.perl}.
As expected, COMET is almost insensitive to this normalization.\footnote{There are exception for the Japanese and Chinese languages for which Moses scripts were not designed to deal with. The scores of the normalized translations should be ignored for these languages.} On the other hand, this normalization has a stronger impact on the BLEU scores, but it can greatly vary from one system to another. For instance for en$\rightarrow$cs, it has no effect on JDExploreAcademy while the score of Online-Y drops by 1.4 BLEU points. For de$\rightarrow$fr, the normalization increases the BLEU score of Online-A by 4.9 point and becomes better than Online-W for which the normalization has no effect on BLEU. Nonetheless, Online-W remains around 10 COMET points better than Online-A.

\section{Evaluation With Different COMET Models}
\label{sec:comet}
In Section \ref{sec:ranking}, we observed that COMET scores can be very high (above 100) or very low (below -100) making their interpretation very difficult. In this section, in an attempt to obtain scores easier to interpret we reevaluate all the translations with 5 different COMET models: 

\begin{itemize}
    \item wmt20-comet-da: The same model used in Section \ref{sec:ranking} trained with direct assessment (DA).
    \item wmt21-comet-da: A more recent model trained with DA on more data.
    \item wmt21-comet-mqm: A model trained with MQM human judgments.
    \item wmt21-comet-qe-mqm: A reference-less model trained with MQM human judgments.
    \item wmt21-comet-qe-da: A reference-less model trained with DA.
\end{itemize}

The results using these 5 models are presented in Section \ref{sec:comettab} from Table \ref{tab:firstcomet} to \ref{tab:lastcomet}.

We can observe that wmt20-comet-da scores are actually quite different from all the other models. While the maximum score obtained by a system with wmt20-comet-da is 104.9 (uk$\rightarrow$cs), the scores obtained with the other 4 models never exceed 15.9 for all translation directions. The absolute scores remain difficult to interpret with the 4 new models. More particularly, with wmt21-comet-da for ja$\rightarrow$en, the best system is scored at 1.1 while for zh$\rightarrow$en, wmt21-comet-da scores are negative for all the systems. We can conclude that absolute COMET scores are not informative whatever model we use. Negative COMET scores can be assigned to an excellent machine translation.

Nonetheless, the rankings obtained by all 5 models are very similar albeit the reference-less models that tend to produce more different rankings.

\section{Conclusion}
\label{sec:ccl}
In this paper, we presented a large-scale evaluation of 185 systems for the 21 translation directions of 11 language pairs. The variety of scenarios, going from high-resource to low-resource language pairs, for closely related or distant language pairs, draws an interesting picture of the state of machine translation. According to three different automatic metrics, significant progress remains to be done to get close to a reference translation, especially for the most difficult scenario involving distant low-resource language pairs.

For only 5 translation directions, WMT22 systems submitted are significantly better than online systems according to COMET. This suggests that the online systems for these translation directions used less training data.

We highlight that while COMET is a state-of-the-art metric for ranking systems, its score are uninformative, or misleading for the readers that are unaware of COMET, as even systems that looks reasonably good may only achieve a negative COMET score. We believe that this peculiarity\footnote{The large majority of NLP evaluation metrics only yield positive scores.} is a major obstacle for its wide adoption in MT research.

We also highlighted that most submitted systems still do not exploit document-level features or context larger than one segment.

It will be interesting to observe how the human evaluation conducted by WMT22 will align with these findings.

\section*{Acknowledgments}
We would like to thank the WMT organizers for releasing the translations and Tom Kocmi for providing preliminary results as well as insightful comments and suggestions on the first draft of this report.

\bibliography{anthology,custom}
\bibliographystyle{acl_natbib}
\clearpage
\appendix

\section{Ranking Tables}
\label{sec:rankingtab}
From Table \ref{tab:firstranking} to \ref{tab:lastranking}.
\begin{table} \definecolor{ashgrey}{rgb}{0.7, 0.75, 0.71}
\scriptsize\begin{subtable}[t]{0.33\textwidth}
\begin{tabular}{rccc}
\toprule
System & Rank & chrF \\
\midrule
\color{ashgrey}Online-W & \color{ashgrey}n/a & \color{ashgrey}79.7 \\
JDExploreAcademy & 1 & 75.0 \\
\color{ashgrey}Lan-Bridge & \color{ashgrey}n/a & \color{ashgrey}74.6 \\
\color{ashgrey}Online-B & \color{ashgrey}n/a & \color{ashgrey}74.5 \\
\color{ashgrey}Online-A & \color{ashgrey}n/a & \color{ashgrey}74.1 \\
CUNI-DocTransformer & 2 & 73.0 \\
CUNI-Transformer & 3 & 72.4 \\
\color{ashgrey}Online-G & \color{ashgrey}n/a & \color{ashgrey}71.1 \\
SHOPLINE-PL & 4 & 70.1 \\
\color{ashgrey}Online-Y & \color{ashgrey}n/a & \color{ashgrey}68.8 \\
ALMAnaCH-Inria & 5 & 58.3 \\
\bottomrule
\end{tabular}
\end{subtable}
\begin{subtable}[t]{0.33\textwidth}
\begin{tabular}{rccc}
\toprule
System & Rank & BLEU \\
\midrule
\color{ashgrey}Online-W & \color{ashgrey}n/a & \color{ashgrey}68.7 \\
JDExploreAcademy & 1 & 60.9 \\
\color{ashgrey}Lan-Bridge & \color{ashgrey}n/a & \color{ashgrey}60.8 \\
\color{ashgrey}Online-B & \color{ashgrey}n/a & \color{ashgrey}60.6 \\
\color{ashgrey}Online-A & \color{ashgrey}n/a & \color{ashgrey}59.4 \\
CUNI-DocTransformer & 2 & 57.9 \\
CUNI-Transformer & 3 & 57.6 \\
\color{ashgrey}Online-G & \color{ashgrey}n/a & \color{ashgrey}54.3 \\
SHOPLINE-PL & 4 & 53.5 \\
\color{ashgrey}Online-Y & \color{ashgrey}n/a & \color{ashgrey}51.0 \\
ALMAnaCH-Inria & 5 & 36.8 \\
\bottomrule
\end{tabular}
\end{subtable}
\begin{subtable}[t]{0.33\textwidth}
\begin{tabular}{rccc}
\toprule
System & Rank & COMET \\
\midrule
\color{ashgrey}Online-W & \color{ashgrey}n/a & \color{ashgrey}77.5 \\
JDExploreAcademy & 1 & 74.8 \\
\color{ashgrey}Online-B & \color{ashgrey}n/a & \color{ashgrey}71.9 \\
\color{ashgrey}Lan-Bridge & \color{ashgrey}n/a & \color{ashgrey}71.8 \\
CUNI-DocTransformer & 2 & 70.6 \\
\color{ashgrey}Online-A & \color{ashgrey}n/a & \color{ashgrey}69.8 \\
CUNI-Transformer & 3 & 69.2 \\
\color{ashgrey}Online-G & \color{ashgrey}n/a & \color{ashgrey}63.0 \\
SHOPLINE-PL & 4 & 61.2 \\
\color{ashgrey}Online-Y & \color{ashgrey}n/a & \color{ashgrey}58.7 \\
ALMAnaCH-Inria & 5 & 19.3 \\
\bottomrule
\end{tabular}
\end{subtable}
\caption{\label{tab:firstranking} Scores for the cs$\rightarrow$en translation task: chrF (nrefs:2|case:mixed|eff:yes|nc:6|nw:0|space:no|version:2.1.0), BLEU (nrefs:2|case:mixed|eff:no|tok:13a|smooth:exp|version:2.1.0), COMET (wmt20-comet-da).}
\end{table}
\begin{table} \definecolor{ashgrey}{rgb}{0.7, 0.75, 0.71}
\scriptsize\begin{subtable}[t]{0.33\textwidth}
\begin{tabular}{rccc}
\toprule
System & Rank & chrF \\
\midrule
\color{ashgrey}Online-B & \color{ashgrey}n/a & \color{ashgrey}71.3 \\
Lan-Bridge & 1 & 70.4 \\
\color{ashgrey}Online-W & \color{ashgrey}n/a & \color{ashgrey}70.4 \\
\color{ashgrey}Online-A & \color{ashgrey}n/a & \color{ashgrey}67.5 \\
JDExploreAcademy & 2 & 67.2 \\
CUNI-DocTransformer & 3 & 66.0 \\
CUNI-Bergamot & 4 & 65.1 \\
\color{ashgrey}Online-G & \color{ashgrey}n/a & \color{ashgrey}64.6 \\
\color{ashgrey}Online-Y & \color{ashgrey}n/a & \color{ashgrey}64.5 \\
CUNI-Transformer & 5 & 64.2 \\
\bottomrule
\end{tabular}
\end{subtable}
\begin{subtable}[t]{0.33\textwidth}
\begin{tabular}{rccc}
\toprule
System & Rank & BLEU \\
\midrule
\color{ashgrey}Online-B & \color{ashgrey}n/a & \color{ashgrey}56.5 \\
Lan-Bridge & 1 & 53.4 \\
\color{ashgrey}Online-W & \color{ashgrey}n/a & \color{ashgrey}53.1 \\
JDExploreAcademy & 2 & 49.4 \\
\color{ashgrey}Online-A & \color{ashgrey}n/a & \color{ashgrey}49.1 \\
CUNI-DocTransformer & 3 & 47.8 \\
CUNI-Bergamot & 4 & 46.2 \\
CUNI-Transformer & 5 & 45.6 \\
\color{ashgrey}Online-G & \color{ashgrey}n/a & \color{ashgrey}44.8 \\
\color{ashgrey}Online-Y & \color{ashgrey}n/a & \color{ashgrey}44.7 \\
\bottomrule
\end{tabular}
\end{subtable}
\begin{subtable}[t]{0.33\textwidth}
\begin{tabular}{rccc}
\toprule
System & Rank & COMET \\
\midrule
\color{ashgrey}Online-W & \color{ashgrey}n/a & \color{ashgrey}97.9 \\
\color{ashgrey}Online-B & \color{ashgrey}n/a & \color{ashgrey}97.5 \\
CUNI-Bergamot & 1 & 96.1 \\
JDExploreAcademy & 2 & 95.3 \\
Lan-Bridge & 3 & 94.7 \\
\color{ashgrey}Online-A & \color{ashgrey}n/a & \color{ashgrey}92.2 \\
CUNI-DocTransformer & 4 & 91.7 \\
CUNI-Transformer & 5 & 86.7 \\
\color{ashgrey}Online-Y & \color{ashgrey}n/a & \color{ashgrey}83.7 \\
\color{ashgrey}Online-G & \color{ashgrey}n/a & \color{ashgrey}82.3 \\
\bottomrule
\end{tabular}
\end{subtable}
\caption{Scores for the en$\rightarrow$cs translation task: chrF (nrefs:2|case:mixed|eff:yes|nc:6|nw:0|space:no|version:2.1.0), BLEU (nrefs:2|case:mixed|eff:no|tok:13a|smooth:exp|version:2.1.0), COMET (wmt20-comet-da).}
\end{table}

\begin{table} \definecolor{ashgrey}{rgb}{0.7, 0.75, 0.71}
\scriptsize\begin{subtable}[t]{0.33\textwidth}
\begin{tabular}{rccc}
\toprule
System & Rank & chrF \\
\midrule
\color{ashgrey}Online-B & \color{ashgrey}n/a & \color{ashgrey}64.0 \\
\color{ashgrey}Lan-Bridge & \color{ashgrey}n/a & \color{ashgrey}64.0 \\
HuaweiTSC & 1 & 62.6 \\
\color{ashgrey}Online-A & \color{ashgrey}n/a & \color{ashgrey}62.2 \\
CUNI-JL-JH & 2 & 61.6 \\
CUNI-Transformer & 3 & 61.6 \\
AMU & 4 & 61.5 \\
\color{ashgrey}CharlesTranslator & \color{ashgrey}n/a & \color{ashgrey}61.5 \\
\color{ashgrey}Online-G & \color{ashgrey}n/a & \color{ashgrey}60.8 \\
\color{ashgrey}Online-Y & \color{ashgrey}n/a & \color{ashgrey}59.6 \\
ALMAnaCH-Inria & 5 & 54.5 \\
\bottomrule
\end{tabular}
\end{subtable}
\begin{subtable}[t]{0.33\textwidth}
\begin{tabular}{rccc}
\toprule
System & Rank & BLEU \\
\midrule
\color{ashgrey}Online-B & \color{ashgrey}n/a & \color{ashgrey}38.3 \\
\color{ashgrey}Lan-Bridge & \color{ashgrey}n/a & \color{ashgrey}38.3 \\
HuaweiTSC & 1 & 36.0 \\
\color{ashgrey}Online-A & \color{ashgrey}n/a & \color{ashgrey}35.9 \\
CUNI-Transformer & 2 & 35.0 \\
CUNI-JL-JH & 3 & 34.8 \\
AMU & 4 & 34.7 \\
\color{ashgrey}CharlesTranslator & \color{ashgrey}n/a & \color{ashgrey}34.3 \\
\color{ashgrey}Online-G & \color{ashgrey}n/a & \color{ashgrey}32.5 \\
\color{ashgrey}Online-Y & \color{ashgrey}n/a & \color{ashgrey}32.1 \\
ALMAnaCH-Inria & 5 & 26.8 \\
\bottomrule
\end{tabular}
\end{subtable}
\begin{subtable}[t]{0.33\textwidth}
\begin{tabular}{rccc}
\toprule
System & Rank & COMET \\
\midrule
AMU & 1 & 99.4 \\
\color{ashgrey}Online-B & \color{ashgrey}n/a & \color{ashgrey}94.3 \\
\color{ashgrey}Lan-Bridge & \color{ashgrey}n/a & \color{ashgrey}91.8 \\
\color{ashgrey}CharlesTranslator & \color{ashgrey}n/a & \color{ashgrey}90.8 \\
HuaweiTSC & 2 & 90.7 \\
CUNI-JL-JH & 3 & 90.1 \\
\color{ashgrey}Online-G & \color{ashgrey}n/a & \color{ashgrey}88.4 \\
\color{ashgrey}Online-A & \color{ashgrey}n/a & \color{ashgrey}87.9 \\
CUNI-Transformer & 4 & 87.4 \\
\color{ashgrey}Online-Y & \color{ashgrey}n/a & \color{ashgrey}78.4 \\
ALMAnaCH-Inria & 5 & 61.4 \\
\bottomrule
\end{tabular}
\end{subtable}
\caption{Scores for the cs$\rightarrow$uk translation task: chrF (nrefs:1|case:mixed|eff:yes|nc:6|nw:0|space:no|version:2.1.0), BLEU (nrefs:1|case:mixed|eff:no|tok:13a|smooth:exp|version:2.1.0), COMET (wmt20-comet-da).}
\end{table}
\begin{table} \definecolor{ashgrey}{rgb}{0.7, 0.75, 0.71}
\scriptsize\begin{subtable}[t]{0.33\textwidth}
\begin{tabular}{rccc}
\toprule
System & Rank & chrF \\
\midrule
AMU & 1 & 60.7 \\
\color{ashgrey}Lan-Bridge & \color{ashgrey}n/a & \color{ashgrey}60.4 \\
\color{ashgrey}Online-B & \color{ashgrey}n/a & \color{ashgrey}60.3 \\
HuaweiTSC & 2 & 59.6 \\
\color{ashgrey}CharlesTranslator & \color{ashgrey}n/a & \color{ashgrey}59.0 \\
CUNI-Transformer & 3 & 59.0 \\
CUNI-JL-JH & 4 & 58.7 \\
\color{ashgrey}Online-A & \color{ashgrey}n/a & \color{ashgrey}57.5 \\
\color{ashgrey}Online-G & \color{ashgrey}n/a & \color{ashgrey}56.3 \\
\color{ashgrey}Online-Y & \color{ashgrey}n/a & \color{ashgrey}55.3 \\
ALMAnaCH-Inria & 5 & 50.7 \\
\bottomrule
\end{tabular}
\end{subtable}
\begin{subtable}[t]{0.33\textwidth}
\begin{tabular}{rccc}
\toprule
System & Rank & BLEU \\
\midrule
AMU & 1 & 37.0 \\
\color{ashgrey}Lan-Bridge & \color{ashgrey}n/a & \color{ashgrey}36.5 \\
\color{ashgrey}Online-B & \color{ashgrey}n/a & \color{ashgrey}36.4 \\
HuaweiTSC & 2 & 36.0 \\
\color{ashgrey}CharlesTranslator & \color{ashgrey}n/a & \color{ashgrey}35.9 \\
CUNI-Transformer & 3 & 35.8 \\
CUNI-JL-JH & 4 & 35.1 \\
\color{ashgrey}Online-A & \color{ashgrey}n/a & \color{ashgrey}33.3 \\
\color{ashgrey}Online-G & \color{ashgrey}n/a & \color{ashgrey}31.5 \\
\color{ashgrey}Online-Y & \color{ashgrey}n/a & \color{ashgrey}29.6 \\
ALMAnaCH-Inria & 5 & 25.3 \\
\bottomrule
\end{tabular}
\end{subtable}
\begin{subtable}[t]{0.33\textwidth}
\begin{tabular}{rccc}
\toprule
System & Rank & COMET \\
\midrule
AMU & 1 & 104.9 \\
\color{ashgrey}Online-B & \color{ashgrey}n/a & \color{ashgrey}96.5 \\
\color{ashgrey}Lan-Bridge & \color{ashgrey}n/a & \color{ashgrey}94.6 \\
HuaweiTSC & 2 & 91.5 \\
\color{ashgrey}CharlesTranslator & \color{ashgrey}n/a & \color{ashgrey}90.2 \\
CUNI-JL-JH & 3 & 89.1 \\
CUNI-Transformer & 4 & 88.5 \\
\color{ashgrey}Online-A & \color{ashgrey}n/a & \color{ashgrey}85.4 \\
\color{ashgrey}Online-G & \color{ashgrey}n/a & \color{ashgrey}84.2 \\
\color{ashgrey}Online-Y & \color{ashgrey}n/a & \color{ashgrey}78.7 \\
ALMAnaCH-Inria & 5 & 62.5 \\
\bottomrule
\end{tabular}
\end{subtable}
\caption{Scores for the uk$\rightarrow$cs translation task: chrF (nrefs:1|case:mixed|eff:yes|nc:6|nw:0|space:no|version:2.1.0), BLEU (nrefs:1|case:mixed|eff:no|tok:13a|smooth:exp|version:2.1.0), COMET (wmt20-comet-da).}
\end{table}

\begin{table} \definecolor{ashgrey}{rgb}{0.7, 0.75, 0.71}
\scriptsize\begin{subtable}[t]{0.33\textwidth}
\begin{tabular}{rccc}
\toprule
System & Rank & chrF \\
\midrule
\color{ashgrey}Online-A & \color{ashgrey}n/a & \color{ashgrey}66.4 \\
\color{ashgrey}Online-G & \color{ashgrey}n/a & \color{ashgrey}66.3 \\
\color{ashgrey}Lan-Bridge & \color{ashgrey}n/a & \color{ashgrey}66.2 \\
JDExploreAcademy & 1 & 65.8 \\
\color{ashgrey}Online-B & \color{ashgrey}n/a & \color{ashgrey}65.7 \\
\color{ashgrey}Online-Y & \color{ashgrey}n/a & \color{ashgrey}65.7 \\
\color{ashgrey}PROMT & \color{ashgrey}n/a & \color{ashgrey}65.6 \\
\color{ashgrey}Online-W & \color{ashgrey}n/a & \color{ashgrey}65.4 \\
LT22 & 2 & 58.4 \\
\bottomrule
\end{tabular}
\end{subtable}
\begin{subtable}[t]{0.33\textwidth}
\begin{tabular}{rccc}
\toprule
System & Rank & BLEU \\
\midrule
\color{ashgrey}Online-A & \color{ashgrey}n/a & \color{ashgrey}50.2 \\
\color{ashgrey}Lan-Bridge & \color{ashgrey}n/a & \color{ashgrey}50.1 \\
\color{ashgrey}Online-B & \color{ashgrey}n/a & \color{ashgrey}49.7 \\
\color{ashgrey}Online-G & \color{ashgrey}n/a & \color{ashgrey}49.7 \\
JDExploreAcademy & 1 & 49.3 \\
\color{ashgrey}Online-Y & \color{ashgrey}n/a & \color{ashgrey}49.3 \\
\color{ashgrey}PROMT & \color{ashgrey}n/a & \color{ashgrey}49.2 \\
\color{ashgrey}Online-W & \color{ashgrey}n/a & \color{ashgrey}48.8 \\
LT22 & 2 & 40.3 \\
\bottomrule
\end{tabular}
\end{subtable}
\begin{subtable}[t]{0.33\textwidth}
\begin{tabular}{rccc}
\toprule
System & Rank & COMET \\
\midrule
JDExploreAcademy & 1 & 58.0 \\
\color{ashgrey}Online-B & \color{ashgrey}n/a & \color{ashgrey}57.0 \\
\color{ashgrey}Lan-Bridge & \color{ashgrey}n/a & \color{ashgrey}56.6 \\
\color{ashgrey}Online-G & \color{ashgrey}n/a & \color{ashgrey}55.3 \\
\color{ashgrey}Online-Y & \color{ashgrey}n/a & \color{ashgrey}54.7 \\
\color{ashgrey}Online-A & \color{ashgrey}n/a & \color{ashgrey}54.5 \\
\color{ashgrey}Online-W & \color{ashgrey}n/a & \color{ashgrey}54.4 \\
\color{ashgrey}PROMT & \color{ashgrey}n/a & \color{ashgrey}51.8 \\
LT22 & 2 & 25.6 \\
\bottomrule
\end{tabular}
\end{subtable}
\caption{Scores for the de$\rightarrow$en translation task: chrF (nrefs:2|case:mixed|eff:yes|nc:6|nw:0|space:no|version:2.1.0), BLEU (nrefs:2|case:mixed|eff:no|tok:13a|smooth:exp|version:2.1.0), COMET (wmt20-comet-da).}
\end{table}
\begin{table} \definecolor{ashgrey}{rgb}{0.7, 0.75, 0.71}
\scriptsize\begin{subtable}[t]{0.33\textwidth}
\begin{tabular}{rccc}
\toprule
System & Rank & chrF \\
\midrule
\color{ashgrey}Online-B & \color{ashgrey}n/a & \color{ashgrey}69.7 \\
JDExploreAcademy & 1 & 69.2 \\
\color{ashgrey}Lan-Bridge & \color{ashgrey}n/a & \color{ashgrey}69.1 \\
\color{ashgrey}Online-A & \color{ashgrey}n/a & \color{ashgrey}68.8 \\
\color{ashgrey}Online-W & \color{ashgrey}n/a & \color{ashgrey}68.7 \\
\color{ashgrey}Online-Y & \color{ashgrey}n/a & \color{ashgrey}68.7 \\
\color{ashgrey}Online-G & \color{ashgrey}n/a & \color{ashgrey}68.4 \\
\color{ashgrey}PROMT & \color{ashgrey}n/a & \color{ashgrey}67.5 \\
\color{ashgrey}OpenNMT & \color{ashgrey}n/a & \color{ashgrey}66.7 \\
\bottomrule
\end{tabular}
\end{subtable}
\begin{subtable}[t]{0.33\textwidth}
\begin{tabular}{rccc}
\toprule
System & Rank & BLEU \\
\midrule
\color{ashgrey}Online-B & \color{ashgrey}n/a & \color{ashgrey}52.3 \\
JDExploreAcademy & 1 & 51.6 \\
\color{ashgrey}Online-Y & \color{ashgrey}n/a & \color{ashgrey}50.5 \\
\color{ashgrey}Online-A & \color{ashgrey}n/a & \color{ashgrey}50.1 \\
\color{ashgrey}Online-G & \color{ashgrey}n/a & \color{ashgrey}49.6 \\
\color{ashgrey}Lan-Bridge & \color{ashgrey}n/a & \color{ashgrey}49.4 \\
\color{ashgrey}PROMT & \color{ashgrey}n/a & \color{ashgrey}49.0 \\
\color{ashgrey}Online-W & \color{ashgrey}n/a & \color{ashgrey}48.9 \\
\color{ashgrey}OpenNMT & \color{ashgrey}n/a & \color{ashgrey}48.4 \\
\bottomrule
\end{tabular}
\end{subtable}
\begin{subtable}[t]{0.33\textwidth}
\begin{tabular}{rccc}
\toprule
System & Rank & COMET \\
\midrule
\color{ashgrey}Online-W & \color{ashgrey}n/a & \color{ashgrey}65.6 \\
JDExploreAcademy & 1 & 63.3 \\
\color{ashgrey}Online-B & \color{ashgrey}n/a & \color{ashgrey}62.3 \\
\color{ashgrey}Online-Y & \color{ashgrey}n/a & \color{ashgrey}61.1 \\
\color{ashgrey}Online-A & \color{ashgrey}n/a & \color{ashgrey}60.6 \\
\color{ashgrey}Online-G & \color{ashgrey}n/a & \color{ashgrey}60.2 \\
\color{ashgrey}Lan-Bridge & \color{ashgrey}n/a & \color{ashgrey}58.8 \\
\color{ashgrey}OpenNMT & \color{ashgrey}n/a & \color{ashgrey}57.3 \\
\color{ashgrey}PROMT & \color{ashgrey}n/a & \color{ashgrey}55.9 \\
\bottomrule
\end{tabular}
\end{subtable}
\caption{Scores for the en$\rightarrow$de translation task: chrF (nrefs:2|case:mixed|eff:yes|nc:6|nw:0|space:no|version:2.1.0), BLEU (nrefs:2|case:mixed|eff:no|tok:13a|smooth:exp|version:2.1.0), COMET (wmt20-comet-da).}
\end{table}

\begin{table} \definecolor{ashgrey}{rgb}{0.7, 0.75, 0.71}
\scriptsize\begin{subtable}[t]{0.33\textwidth}
\begin{tabular}{rccc}
\toprule
System & Rank & chrF \\
\midrule
\color{ashgrey}Online-B & \color{ashgrey}n/a & \color{ashgrey}74.6 \\
\color{ashgrey}Online-Y & \color{ashgrey}n/a & \color{ashgrey}66.8 \\
\color{ashgrey}Online-W & \color{ashgrey}n/a & \color{ashgrey}65.5 \\
\color{ashgrey}Online-A & \color{ashgrey}n/a & \color{ashgrey}64.5 \\
\color{ashgrey}Online-G & \color{ashgrey}n/a & \color{ashgrey}62.7 \\
LT22 & 1 & 54.4 \\
\bottomrule
\end{tabular}
\end{subtable}
\begin{subtable}[t]{0.33\textwidth}
\begin{tabular}{rccc}
\toprule
System & Rank & BLEU \\
\midrule
\color{ashgrey}Online-B & \color{ashgrey}n/a & \color{ashgrey}58.4 \\
\color{ashgrey}Online-Y & \color{ashgrey}n/a & \color{ashgrey}46.2 \\
\color{ashgrey}Online-W & \color{ashgrey}n/a & \color{ashgrey}43.6 \\
\color{ashgrey}Online-A & \color{ashgrey}n/a & \color{ashgrey}41.3 \\
\color{ashgrey}Online-G & \color{ashgrey}n/a & \color{ashgrey}39.0 \\
LT22 & 1 & 28.3 \\
\bottomrule
\end{tabular}
\end{subtable}
\begin{subtable}[t]{0.33\textwidth}
\begin{tabular}{rccc}
\toprule
System & Rank & COMET \\
\midrule
\color{ashgrey}Online-B & \color{ashgrey}n/a & \color{ashgrey}70.5 \\
\color{ashgrey}Online-W & \color{ashgrey}n/a & \color{ashgrey}63.7 \\
\color{ashgrey}Online-Y & \color{ashgrey}n/a & \color{ashgrey}57.8 \\
\color{ashgrey}Online-A & \color{ashgrey}n/a & \color{ashgrey}52.3 \\
\color{ashgrey}Online-G & \color{ashgrey}n/a & \color{ashgrey}44.9 \\
LT22 & 1 & 10.5 \\
\bottomrule
\end{tabular}
\end{subtable}
\caption{Scores for the de$\rightarrow$fr translation task: chrF (nrefs:1|case:mixed|eff:yes|nc:6|nw:0|space:no|version:2.1.0), BLEU (nrefs:1|case:mixed|eff:no|tok:13a|smooth:exp|version:2.1.0), COMET (wmt20-comet-da).}
\end{table}
\begin{table} \definecolor{ashgrey}{rgb}{0.7, 0.75, 0.71}
\scriptsize\begin{subtable}[t]{0.33\textwidth}
\begin{tabular}{rccc}
\toprule
System & Rank & chrF \\
\midrule
\color{ashgrey}Online-W & \color{ashgrey}n/a & \color{ashgrey}81.2 \\
\color{ashgrey}Online-B & \color{ashgrey}n/a & \color{ashgrey}68.7 \\
eTranslation & 1 & 68.4 \\
\color{ashgrey}Online-Y & \color{ashgrey}n/a & \color{ashgrey}67.5 \\
\color{ashgrey}Online-A & \color{ashgrey}n/a & \color{ashgrey}67.2 \\
\color{ashgrey}Online-G & \color{ashgrey}n/a & \color{ashgrey}66.0 \\
\color{ashgrey}Lan-Bridge & \color{ashgrey}n/a & \color{ashgrey}65.0 \\
\bottomrule
\end{tabular}
\end{subtable}
\begin{subtable}[t]{0.33\textwidth}
\begin{tabular}{rccc}
\toprule
System & Rank & BLEU \\
\midrule
\color{ashgrey}Online-W & \color{ashgrey}n/a & \color{ashgrey}64.8 \\
\color{ashgrey}Online-B & \color{ashgrey}n/a & \color{ashgrey}46.6 \\
eTranslation & 1 & 46.5 \\
\color{ashgrey}Online-Y & \color{ashgrey}n/a & \color{ashgrey}45.0 \\
\color{ashgrey}Online-A & \color{ashgrey}n/a & \color{ashgrey}44.4 \\
\color{ashgrey}Lan-Bridge & \color{ashgrey}n/a & \color{ashgrey}41.8 \\
\color{ashgrey}Online-G & \color{ashgrey}n/a & \color{ashgrey}41.1 \\
\bottomrule
\end{tabular}
\end{subtable}
\begin{subtable}[t]{0.33\textwidth}
\begin{tabular}{rccc}
\toprule
System & Rank & COMET \\
\midrule
\color{ashgrey}Online-W & \color{ashgrey}n/a & \color{ashgrey}77.9 \\
\color{ashgrey}Online-B & \color{ashgrey}n/a & \color{ashgrey}63.8 \\
\color{ashgrey}Online-Y & \color{ashgrey}n/a & \color{ashgrey}61.6 \\
\color{ashgrey}Online-A & \color{ashgrey}n/a & \color{ashgrey}59.3 \\
eTranslation & 1 & 55.5 \\
\color{ashgrey}Lan-Bridge & \color{ashgrey}n/a & \color{ashgrey}51.2 \\
\color{ashgrey}Online-G & \color{ashgrey}n/a & \color{ashgrey}48.3 \\
\bottomrule
\end{tabular}
\end{subtable}
\caption{Scores for the fr$\rightarrow$de translation task: chrF (nrefs:1|case:mixed|eff:yes|nc:6|nw:0|space:no|version:2.1.0), BLEU (nrefs:1|case:mixed|eff:no|tok:13a|smooth:exp|version:2.1.0), COMET (wmt20-comet-da).}
\end{table}

\begin{table} \definecolor{ashgrey}{rgb}{0.7, 0.75, 0.71}
\scriptsize\begin{subtable}[t]{0.33\textwidth}
\begin{tabular}{rccc}
\toprule
System & Rank & chrF \\
\midrule
\color{ashgrey}Online-B & \color{ashgrey}n/a & \color{ashgrey}58.5 \\
\color{ashgrey}Lan-Bridge & \color{ashgrey}n/a & \color{ashgrey}58.5 \\
HuaweiTSC & 1 & 56.8 \\
\color{ashgrey}Online-A & \color{ashgrey}n/a & \color{ashgrey}56.5 \\
SRPOL & 2 & 56.3 \\
NiuTrans & 3 & 56.3 \\
\color{ashgrey}Online-Y & \color{ashgrey}n/a & \color{ashgrey}54.3 \\
\color{ashgrey}Online-G & \color{ashgrey}n/a & \color{ashgrey}53.2 \\
\bottomrule
\end{tabular}
\end{subtable}
\begin{subtable}[t]{0.33\textwidth}
\begin{tabular}{rccc}
\toprule
System & Rank & BLEU \\
\midrule
\color{ashgrey}Online-B & \color{ashgrey}n/a & \color{ashgrey}31.5 \\
\color{ashgrey}Lan-Bridge & \color{ashgrey}n/a & \color{ashgrey}31.5 \\
HuaweiTSC & 1 & 29.9 \\
NiuTrans & 2 & 29.3 \\
\color{ashgrey}Online-A & \color{ashgrey}n/a & \color{ashgrey}29.1 \\
SRPOL & 3 & 29.1 \\
\color{ashgrey}Online-Y & \color{ashgrey}n/a & \color{ashgrey}26.6 \\
\color{ashgrey}Online-G & \color{ashgrey}n/a & \color{ashgrey}25.7 \\
\bottomrule
\end{tabular}
\end{subtable}
\begin{subtable}[t]{0.33\textwidth}
\begin{tabular}{rccc}
\toprule
System & Rank & COMET \\
\midrule
\color{ashgrey}Online-B & \color{ashgrey}n/a & \color{ashgrey}80.4 \\
\color{ashgrey}Lan-Bridge & \color{ashgrey}n/a & \color{ashgrey}79.6 \\
\color{ashgrey}Online-A & \color{ashgrey}n/a & \color{ashgrey}69.5 \\
SRPOL & 1 & 69.5 \\
HuaweiTSC & 2 & 67.7 \\
NiuTrans & 3 & 65.6 \\
\color{ashgrey}Online-G & \color{ashgrey}n/a & \color{ashgrey}64.3 \\
\color{ashgrey}Online-Y & \color{ashgrey}n/a & \color{ashgrey}56.8 \\
\bottomrule
\end{tabular}
\end{subtable}
\caption{Scores for the en$\rightarrow$hr translation task: chrF (nrefs:1|case:mixed|eff:yes|nc:6|nw:0|space:no|version:2.1.0), BLEU (nrefs:1|case:mixed|eff:no|tok:13a|smooth:exp|version:2.1.0), COMET (wmt20-comet-da).}
\end{table}

\begin{table} \definecolor{ashgrey}{rgb}{0.7, 0.75, 0.71}
\scriptsize\begin{subtable}[t]{0.33\textwidth}
\begin{tabular}{rccc}
\toprule
System & Rank & chrF \\
\midrule
\color{ashgrey}Online-W & \color{ashgrey}n/a & \color{ashgrey}51.7 \\
NT5 & 1 & 51.3 \\
JDExploreAcademy & 2 & 50.1 \\
\color{ashgrey}Online-B & \color{ashgrey}n/a & \color{ashgrey}49.9 \\
DLUT & 3 & 49.8 \\
\color{ashgrey}LanguageX & \color{ashgrey}n/a & \color{ashgrey}49.1 \\
\color{ashgrey}Lan-Bridge & \color{ashgrey}n/a & \color{ashgrey}48.7 \\
\color{ashgrey}Online-A & \color{ashgrey}n/a & \color{ashgrey}48.4 \\
NAIST-NICT-TIT & 4 & 48.3 \\
\color{ashgrey}Online-Y & \color{ashgrey}n/a & \color{ashgrey}48.2 \\
AISP-SJTU & 5 & 48.0 \\
\color{ashgrey}Online-G & \color{ashgrey}n/a & \color{ashgrey}45.7 \\
KYB & 6 & 43.4 \\
\color{ashgrey}AIST & \color{ashgrey}n/a & \color{ashgrey}11.4 \\
\bottomrule
\end{tabular}
\end{subtable}
\begin{subtable}[t]{0.33\textwidth}
\begin{tabular}{rccc}
\toprule
System & Rank & BLEU \\
\midrule
\color{ashgrey}Online-W & \color{ashgrey}n/a & \color{ashgrey}27.8 \\
NT5 & 1 & 26.6 \\
JDExploreAcademy & 2 & 25.6 \\
DLUT & 3 & 24.8 \\
\color{ashgrey}Online-B & \color{ashgrey}n/a & \color{ashgrey}24.7 \\
\color{ashgrey}Online-A & \color{ashgrey}n/a & \color{ashgrey}22.8 \\
\color{ashgrey}Lan-Bridge & \color{ashgrey}n/a & \color{ashgrey}22.8 \\
NAIST-NICT-TIT & 4 & 22.7 \\
\color{ashgrey}LanguageX & \color{ashgrey}n/a & \color{ashgrey}22.4 \\
AISP-SJTU & 5 & 22.0 \\
\color{ashgrey}Online-Y & \color{ashgrey}n/a & \color{ashgrey}21.5 \\
\color{ashgrey}Online-G & \color{ashgrey}n/a & \color{ashgrey}19.7 \\
KYB & 6 & 18.1 \\
\color{ashgrey}AIST & \color{ashgrey}n/a & \color{ashgrey}0.1 \\
\bottomrule
\end{tabular}
\end{subtable}
\begin{subtable}[t]{0.33\textwidth}
\begin{tabular}{rccc}
\toprule
System & Rank & COMET \\
\midrule
NT5 & 1 & 42.1 \\
\color{ashgrey}Online-W & \color{ashgrey}n/a & \color{ashgrey}41.2 \\
JDExploreAcademy & 2 & 40.6 \\
\color{ashgrey}Online-B & \color{ashgrey}n/a & \color{ashgrey}39.7 \\
DLUT & 3 & 37.3 \\
NAIST-NICT-TIT & 4 & 33.4 \\
\color{ashgrey}LanguageX & \color{ashgrey}n/a & \color{ashgrey}33.0 \\
\color{ashgrey}Online-A & \color{ashgrey}n/a & \color{ashgrey}32.9 \\
\color{ashgrey}Online-Y & \color{ashgrey}n/a & \color{ashgrey}32.4 \\
\color{ashgrey}Lan-Bridge & \color{ashgrey}n/a & \color{ashgrey}32.0 \\
AISP-SJTU & 5 & 30.2 \\
\color{ashgrey}Online-G & \color{ashgrey}n/a & \color{ashgrey}22.3 \\
KYB & 6 & 17.4 \\
\color{ashgrey}AIST & \color{ashgrey}n/a & \color{ashgrey}-152.7 \\
\bottomrule
\end{tabular}
\end{subtable}
\caption{Scores for the ja$\rightarrow$en translation task: chrF (nrefs:1|case:mixed|eff:yes|nc:6|nw:0|space:no|version:2.1.0), BLEU (nrefs:1|case:mixed|eff:no|tok:13a|smooth:exp|version:2.1.0), COMET (wmt20-comet-da).}
\end{table}
\begin{table} \definecolor{ashgrey}{rgb}{0.7, 0.75, 0.71}
\scriptsize\begin{subtable}[t]{0.33\textwidth}
\begin{tabular}{rccc}
\toprule
System & Rank & chrF \\
\midrule
NT5 & 1 & 36.8 \\
JDExploreAcademy & 2 & 36.1 \\
\color{ashgrey}LanguageX & \color{ashgrey}n/a & \color{ashgrey}36.1 \\
DLUT & 3 & 36.1 \\
\color{ashgrey}Online-B & \color{ashgrey}n/a & \color{ashgrey}35.5 \\
\color{ashgrey}Online-W & \color{ashgrey}n/a & \color{ashgrey}35.2 \\
\color{ashgrey}Online-Y & \color{ashgrey}n/a & \color{ashgrey}34.4 \\
\color{ashgrey}Lan-Bridge & \color{ashgrey}n/a & \color{ashgrey}34.1 \\
\color{ashgrey}Online-A & \color{ashgrey}n/a & \color{ashgrey}34.1 \\
AISP-SJTU & 4 & 33.9 \\
NAIST-NICT-TIT & 5 & 33.8 \\
KYB & 6 & 28.6 \\
\color{ashgrey}Online-G & \color{ashgrey}n/a & \color{ashgrey}28.0 \\
\bottomrule
\end{tabular}
\end{subtable}
\begin{subtable}[t]{0.33\textwidth}
\begin{tabular}{rccc}
\toprule
System & Rank & BLEU \\
\midrule
NT5 & 1 & 27.6 \\
\color{ashgrey}LanguageX & \color{ashgrey}n/a & \color{ashgrey}27.2 \\
DLUT & 2 & 27.2 \\
\color{ashgrey}Online-W & \color{ashgrey}n/a & \color{ashgrey}27.1 \\
JDExploreAcademy & 3 & 27.0 \\
\color{ashgrey}Online-B & \color{ashgrey}n/a & \color{ashgrey}26.5 \\
\color{ashgrey}Online-Y & \color{ashgrey}n/a & \color{ashgrey}25.2 \\
AISP-SJTU & 4 & 25.0 \\
\color{ashgrey}Lan-Bridge & \color{ashgrey}n/a & \color{ashgrey}24.8 \\
NAIST-NICT-TIT & 5 & 24.6 \\
\color{ashgrey}Online-A & \color{ashgrey}n/a & \color{ashgrey}24.0 \\
KYB & 6 & 18.9 \\
\color{ashgrey}Online-G & \color{ashgrey}n/a & \color{ashgrey}18.6 \\
\bottomrule
\end{tabular}
\end{subtable}
\begin{subtable}[t]{0.33\textwidth}
\begin{tabular}{rccc}
\toprule
System & Rank & COMET \\
\midrule
JDExploreAcademy & 1 & 65.2 \\
NT5 & 2 & 64.1 \\
\color{ashgrey}LanguageX & \color{ashgrey}n/a & \color{ashgrey}62.1 \\
\color{ashgrey}Online-B & \color{ashgrey}n/a & \color{ashgrey}60.8 \\
DLUT & 3 & 60.5 \\
\color{ashgrey}Online-W & \color{ashgrey}n/a & \color{ashgrey}59.8 \\
\color{ashgrey}Online-Y & \color{ashgrey}n/a & \color{ashgrey}56.9 \\
\color{ashgrey}Lan-Bridge & \color{ashgrey}n/a & \color{ashgrey}56.6 \\
\color{ashgrey}Online-A & \color{ashgrey}n/a & \color{ashgrey}53.7 \\
NAIST-NICT-TIT & 4 & 53.4 \\
AISP-SJTU & 5 & 52.4 \\
KYB & 6 & 31.8 \\
\color{ashgrey}Online-G & \color{ashgrey}n/a & \color{ashgrey}25.0 \\
\bottomrule
\end{tabular}
\end{subtable}
\caption{Scores for the en$\rightarrow$ja translation task: chrF (nrefs:1|case:mixed|eff:yes|nc:6|nw:0|space:no|version:2.1.0), BLEU (nrefs:1|case:mixed|eff:no|tok:ja-mecab-0.996-IPA|smooth:exp|version:2.1.0), COMET (wmt20-comet-da).}
\end{table}

\begin{table} \definecolor{ashgrey}{rgb}{0.7, 0.75, 0.71}
\scriptsize\begin{subtable}[t]{0.33\textwidth}
\begin{tabular}{rccc}
\toprule
System & Rank & chrF \\
\midrule
\color{ashgrey}TartuNLP & \color{ashgrey}n/a & \color{ashgrey}53.5 \\
\color{ashgrey}TAL-SJTU & \color{ashgrey}n/a & \color{ashgrey}53.2 \\
\color{ashgrey}HuaweiTSC & \color{ashgrey}n/a & \color{ashgrey}48.4 \\
\color{ashgrey}Liv4ever & \color{ashgrey}n/a & \color{ashgrey}46.7 \\
NiuTrans & 1 & 35.6 \\
\bottomrule
\end{tabular}
\end{subtable}
\begin{subtable}[t]{0.33\textwidth}
\begin{tabular}{rccc}
\toprule
System & Rank & BLEU \\
\midrule
\color{ashgrey}TAL-SJTU & \color{ashgrey}n/a & \color{ashgrey}30.4 \\
\color{ashgrey}TartuNLP & \color{ashgrey}n/a & \color{ashgrey}29.9 \\
\color{ashgrey}HuaweiTSC & \color{ashgrey}n/a & \color{ashgrey}23.4 \\
\color{ashgrey}Liv4ever & \color{ashgrey}n/a & \color{ashgrey}23.3 \\
NiuTrans & 1 & 13.0 \\
\bottomrule
\end{tabular}
\end{subtable}
\begin{subtable}[t]{0.33\textwidth}
\begin{tabular}{rccc}
\toprule
System & Rank & COMET \\
\midrule
\color{ashgrey}TartuNLP & \color{ashgrey}n/a & \color{ashgrey}-5.7 \\
\color{ashgrey}TAL-SJTU & \color{ashgrey}n/a & \color{ashgrey}-8.3 \\
\color{ashgrey}HuaweiTSC & \color{ashgrey}n/a & \color{ashgrey}-27.2 \\
\color{ashgrey}Liv4ever & \color{ashgrey}n/a & \color{ashgrey}-44.0 \\
NiuTrans & 1 & -88.3 \\
\bottomrule
\end{tabular}
\end{subtable}
\caption{Scores for the liv$\rightarrow$en translation task: chrF (nrefs:1|case:mixed|eff:yes|nc:6|nw:0|space:no|version:2.1.0), BLEU (nrefs:1|case:mixed|eff:no|tok:13a|smooth:exp|version:2.1.0), COMET (wmt20-comet-da).}
\end{table}

\begin{table} \definecolor{ashgrey}{rgb}{0.7, 0.75, 0.71}
\scriptsize\begin{subtable}[t]{0.33\textwidth}
\begin{tabular}{rccc}
\toprule
System & Rank & chrF \\
\midrule
\color{ashgrey}TAL-SJTU & \color{ashgrey}n/a & \color{ashgrey}43.8 \\
\color{ashgrey}Liv4ever & \color{ashgrey}n/a & \color{ashgrey}39.6 \\
\color{ashgrey}TartuNLP & \color{ashgrey}n/a & \color{ashgrey}39.2 \\
\color{ashgrey}HuaweiTSC & \color{ashgrey}n/a & \color{ashgrey}37.7 \\
NiuTrans & 1 & 30.5 \\
\bottomrule
\end{tabular}
\end{subtable}
\begin{subtable}[t]{0.33\textwidth}
\begin{tabular}{rccc}
\toprule
System & Rank & BLEU \\
\midrule
\color{ashgrey}TAL-SJTU & \color{ashgrey}n/a & \color{ashgrey}17.0 \\
\color{ashgrey}TartuNLP & \color{ashgrey}n/a & \color{ashgrey}15.0 \\
\color{ashgrey}Liv4ever & \color{ashgrey}n/a & \color{ashgrey}14.7 \\
\color{ashgrey}HuaweiTSC & \color{ashgrey}n/a & \color{ashgrey}12.8 \\
NiuTrans & 1 & 12.3 \\
\bottomrule
\end{tabular}
\end{subtable}
\begin{subtable}[t]{0.33\textwidth}
\begin{tabular}{rccc}
\toprule
System & Rank & COMET \\
\midrule
\color{ashgrey}TAL-SJTU & \color{ashgrey}n/a & \color{ashgrey}-29.5 \\
\color{ashgrey}TartuNLP & \color{ashgrey}n/a & \color{ashgrey}-36.8 \\
\color{ashgrey}HuaweiTSC & \color{ashgrey}n/a & \color{ashgrey}-38.8 \\
\color{ashgrey}Liv4ever & \color{ashgrey}n/a & \color{ashgrey}-39.4 \\
NiuTrans & 1 & -81.9 \\
\bottomrule
\end{tabular}
\end{subtable}
\caption{Scores for the en$\rightarrow$liv translation task: chrF (nrefs:1|case:mixed|eff:yes|nc:6|nw:0|space:no|version:2.1.0), BLEU (nrefs:1|case:mixed|eff:no|tok:13a|smooth:exp|version:2.1.0), COMET (wmt20-comet-da).}
\end{table}

\begin{table} \definecolor{ashgrey}{rgb}{0.7, 0.75, 0.71}
\scriptsize\begin{subtable}[t]{0.33\textwidth}
\begin{tabular}{rccc}
\toprule
System & Rank & chrF \\
\midrule
\color{ashgrey}Online-G & \color{ashgrey}n/a & \color{ashgrey}70.0 \\
JDExploreAcademy & 1 & 68.9 \\
\color{ashgrey}Lan-Bridge & \color{ashgrey}n/a & \color{ashgrey}68.5 \\
HuaweiTSC & 2 & 68.5 \\
\color{ashgrey}Online-B & \color{ashgrey}n/a & \color{ashgrey}68.3 \\
\color{ashgrey}Online-A & \color{ashgrey}n/a & \color{ashgrey}68.3 \\
\color{ashgrey}Online-Y & \color{ashgrey}n/a & \color{ashgrey}68.2 \\
SRPOL & 3 & 67.2 \\
\color{ashgrey}Online-W & \color{ashgrey}n/a & \color{ashgrey}66.3 \\
ALMAnaCH-Inria & 4 & 57.9 \\
\bottomrule
\end{tabular}
\end{subtable}
\begin{subtable}[t]{0.33\textwidth}
\begin{tabular}{rccc}
\toprule
System & Rank & BLEU \\
\midrule
\color{ashgrey}Online-G & \color{ashgrey}n/a & \color{ashgrey}46.7 \\
\color{ashgrey}Lan-Bridge & \color{ashgrey}n/a & \color{ashgrey}45.2 \\
JDExploreAcademy & 1 & 45.1 \\
HuaweiTSC & 2 & 45.1 \\
\color{ashgrey}Online-B & \color{ashgrey}n/a & \color{ashgrey}45.0 \\
\color{ashgrey}Online-A & \color{ashgrey}n/a & \color{ashgrey}43.9 \\
\color{ashgrey}Online-Y & \color{ashgrey}n/a & \color{ashgrey}43.8 \\
SRPOL & 3 & 43.6 \\
\color{ashgrey}Online-W & \color{ashgrey}n/a & \color{ashgrey}42.6 \\
ALMAnaCH-Inria & 4 & 30.3 \\
\bottomrule
\end{tabular}
\end{subtable}
\begin{subtable}[t]{0.33\textwidth}
\begin{tabular}{rccc}
\toprule
System & Rank & COMET \\
\midrule
\color{ashgrey}Online-G & \color{ashgrey}n/a & \color{ashgrey}65.2 \\
JDExploreAcademy & 1 & 64.9 \\
\color{ashgrey}Online-Y & \color{ashgrey}n/a & \color{ashgrey}64.1 \\
\color{ashgrey}Lan-Bridge & \color{ashgrey}n/a & \color{ashgrey}63.1 \\
\color{ashgrey}Online-B & \color{ashgrey}n/a & \color{ashgrey}63.1 \\
\color{ashgrey}Online-A & \color{ashgrey}n/a & \color{ashgrey}62.2 \\
\color{ashgrey}Online-W & \color{ashgrey}n/a & \color{ashgrey}61.6 \\
HuaweiTSC & 2 & 60.9 \\
SRPOL & 3 & 59.5 \\
ALMAnaCH-Inria & 4 & 26.9 \\
\bottomrule
\end{tabular}
\end{subtable}
\caption{Scores for the ru$\rightarrow$en translation task: chrF (nrefs:1|case:mixed|eff:yes|nc:6|nw:0|space:no|version:2.1.0), BLEU (nrefs:1|case:mixed|eff:no|tok:13a|smooth:exp|version:2.1.0), COMET (wmt20-comet-da).}
\end{table}
\begin{table} \definecolor{ashgrey}{rgb}{0.7, 0.75, 0.71}
\scriptsize\begin{subtable}[t]{0.33\textwidth}
\begin{tabular}{rccc}
\toprule
System & Rank & chrF \\
\midrule
\color{ashgrey}Online-B & \color{ashgrey}n/a & \color{ashgrey}59.7 \\
\color{ashgrey}Online-G & \color{ashgrey}n/a & \color{ashgrey}59.5 \\
\color{ashgrey}Lan-Bridge & \color{ashgrey}n/a & \color{ashgrey}59.0 \\
JDExploreAcademy & 1 & 58.4 \\
\color{ashgrey}Online-W & \color{ashgrey}n/a & \color{ashgrey}58.3 \\
\color{ashgrey}Online-Y & \color{ashgrey}n/a & \color{ashgrey}58.3 \\
\color{ashgrey}Online-A & \color{ashgrey}n/a & \color{ashgrey}58.1 \\
SRPOL & 2 & 56.4 \\
\color{ashgrey}PROMT & \color{ashgrey}n/a & \color{ashgrey}56.1 \\
HuaweiTSC & 3 & 56.1 \\
eTranslation & 4 & 55.8 \\
\bottomrule
\end{tabular}
\end{subtable}
\begin{subtable}[t]{0.33\textwidth}
\begin{tabular}{rccc}
\toprule
System & Rank & BLEU \\
\midrule
\color{ashgrey}Online-B & \color{ashgrey}n/a & \color{ashgrey}34.9 \\
\color{ashgrey}Online-Y & \color{ashgrey}n/a & \color{ashgrey}33.2 \\
\color{ashgrey}Online-A & \color{ashgrey}n/a & \color{ashgrey}33.1 \\
\color{ashgrey}Online-G & \color{ashgrey}n/a & \color{ashgrey}32.8 \\
JDExploreAcademy & 1 & 32.7 \\
\color{ashgrey}Lan-Bridge & \color{ashgrey}n/a & \color{ashgrey}32.6 \\
\color{ashgrey}Online-W & \color{ashgrey}n/a & \color{ashgrey}32.4 \\
HuaweiTSC & 2 & 30.8 \\
\color{ashgrey}PROMT & \color{ashgrey}n/a & \color{ashgrey}30.6 \\
SRPOL & 3 & 30.4 \\
eTranslation & 4 & 29.8 \\
\bottomrule
\end{tabular}
\end{subtable}
\begin{subtable}[t]{0.33\textwidth}
\begin{tabular}{rccc}
\toprule
System & Rank & COMET \\
\midrule
\color{ashgrey}Online-W & \color{ashgrey}n/a & \color{ashgrey}75.1 \\
\color{ashgrey}Online-G & \color{ashgrey}n/a & \color{ashgrey}73.2 \\
\color{ashgrey}Online-B & \color{ashgrey}n/a & \color{ashgrey}72.9 \\
\color{ashgrey}Online-Y & \color{ashgrey}n/a & \color{ashgrey}69.8 \\
JDExploreAcademy & 1 & 69.7 \\
\color{ashgrey}Lan-Bridge & \color{ashgrey}n/a & \color{ashgrey}67.4 \\
\color{ashgrey}Online-A & \color{ashgrey}n/a & \color{ashgrey}67.4 \\
\color{ashgrey}PROMT & \color{ashgrey}n/a & \color{ashgrey}60.4 \\
SRPOL & 2 & 59.8 \\
HuaweiTSC & 3 & 59.2 \\
eTranslation & 4 & 58.0 \\
\bottomrule
\end{tabular}
\end{subtable}
\caption{Scores for the en$\rightarrow$ru translation task: chrF (nrefs:1|case:mixed|eff:yes|nc:6|nw:0|space:no|version:2.1.0), BLEU (nrefs:1|case:mixed|eff:no|tok:13a|smooth:exp|version:2.1.0), COMET (wmt20-comet-da).}
\end{table}

\begin{table} \definecolor{ashgrey}{rgb}{0.7, 0.75, 0.71}
\scriptsize\begin{subtable}[t]{0.33\textwidth}
\begin{tabular}{rccc}
\toprule
System & Rank & chrF \\
\midrule
\color{ashgrey}Online-G & \color{ashgrey}n/a & \color{ashgrey}55.5 \\
Lan-Bridge & 1 & 28.3 \\
\bottomrule
\end{tabular}
\end{subtable}
\begin{subtable}[t]{0.33\textwidth}
\begin{tabular}{rccc}
\toprule
System & Rank & BLEU \\
\midrule
\color{ashgrey}Online-G & \color{ashgrey}n/a & \color{ashgrey}29.6 \\
Lan-Bridge & 1 & 7.1 \\
\bottomrule
\end{tabular}
\end{subtable}
\begin{subtable}[t]{0.33\textwidth}
\begin{tabular}{rccc}
\toprule
System & Rank & COMET \\
\midrule
\color{ashgrey}Online-G & \color{ashgrey}n/a & \color{ashgrey}31.1 \\
Lan-Bridge & 1 & -75.9 \\
\bottomrule
\end{tabular}
\end{subtable}
\caption{Scores for the sah$\rightarrow$ru translation task: chrF (nrefs:1|case:mixed|eff:yes|nc:6|nw:0|space:no|version:2.1.0), BLEU (nrefs:1|case:mixed|eff:no|tok:13a|smooth:exp|version:2.1.0), COMET (wmt20-comet-da).}
\end{table}

\begin{table} \definecolor{ashgrey}{rgb}{0.7, 0.75, 0.71}
\scriptsize\begin{subtable}[t]{0.33\textwidth}
\begin{tabular}{rccc}
\toprule
System & Rank & chrF \\
\midrule
Lan-Bridge & 1 & 56.6 \\
\color{ashgrey}Online-G & \color{ashgrey}n/a & \color{ashgrey}47.0 \\
\bottomrule
\end{tabular}
\end{subtable}
\begin{subtable}[t]{0.33\textwidth}
\begin{tabular}{rccc}
\toprule
System & Rank & BLEU \\
\midrule
Lan-Bridge & 1 & 15.3 \\
\color{ashgrey}Online-G & \color{ashgrey}n/a & \color{ashgrey}14.7 \\
\bottomrule
\end{tabular}
\end{subtable}
\begin{subtable}[t]{0.33\textwidth}
\begin{tabular}{rccc}
\toprule
System & Rank & COMET \\
\midrule
\color{ashgrey}Online-G & \color{ashgrey}n/a & \color{ashgrey}-17.1 \\
Lan-Bridge & 1 & -48.0 \\
\bottomrule
\end{tabular}
\end{subtable}
\caption{Scores for the ru$\rightarrow$sah translation task: chrF (nrefs:1|case:mixed|eff:yes|nc:6|nw:0|space:no|version:2.1.0), BLEU (nrefs:1|case:mixed|eff:no|tok:13a|smooth:exp|version:2.1.0), COMET (wmt20-comet-da).}
\end{table}

\begin{table} \definecolor{ashgrey}{rgb}{0.7, 0.75, 0.71}
\scriptsize\begin{subtable}[t]{0.33\textwidth}
\begin{tabular}{rccc}
\toprule
System & Rank & chrF \\
\midrule
\color{ashgrey}Lan-Bridge & \color{ashgrey}n/a & \color{ashgrey}67.3 \\
\color{ashgrey}Online-B & \color{ashgrey}n/a & \color{ashgrey}67.2 \\
\color{ashgrey}Online-G & \color{ashgrey}n/a & \color{ashgrey}66.0 \\
\color{ashgrey}Online-A & \color{ashgrey}n/a & \color{ashgrey}65.2 \\
\color{ashgrey}PROMT & \color{ashgrey}n/a & \color{ashgrey}64.7 \\
\color{ashgrey}Online-Y & \color{ashgrey}n/a & \color{ashgrey}64.6 \\
\color{ashgrey}ARC-NKUA & \color{ashgrey}n/a & \color{ashgrey}64.6 \\
HuaweiTSC & 1 & 63.9 \\
ALMAnaCH-Inria & 2 & 55.6 \\
\bottomrule
\end{tabular}
\end{subtable}
\begin{subtable}[t]{0.33\textwidth}
\begin{tabular}{rccc}
\toprule
System & Rank & BLEU \\
\midrule
\color{ashgrey}Lan-Bridge & \color{ashgrey}n/a & \color{ashgrey}44.6 \\
\color{ashgrey}Online-B & \color{ashgrey}n/a & \color{ashgrey}44.4 \\
\color{ashgrey}Online-G & \color{ashgrey}n/a & \color{ashgrey}43.2 \\
\color{ashgrey}Online-A & \color{ashgrey}n/a & \color{ashgrey}42.3 \\
\color{ashgrey}PROMT & \color{ashgrey}n/a & \color{ashgrey}42.1 \\
\color{ashgrey}ARC-NKUA & \color{ashgrey}n/a & \color{ashgrey}41.9 \\
\color{ashgrey}Online-Y & \color{ashgrey}n/a & \color{ashgrey}41.8 \\
HuaweiTSC & 1 & 41.6 \\
ALMAnaCH-Inria & 2 & 30.0 \\
\bottomrule
\end{tabular}
\end{subtable}
\begin{subtable}[t]{0.33\textwidth}
\begin{tabular}{rccc}
\toprule
System & Rank & COMET \\
\midrule
\color{ashgrey}Online-B & \color{ashgrey}n/a & \color{ashgrey}62.6 \\
\color{ashgrey}Lan-Bridge & \color{ashgrey}n/a & \color{ashgrey}62.4 \\
\color{ashgrey}Online-G & \color{ashgrey}n/a & \color{ashgrey}57.5 \\
\color{ashgrey}Online-A & \color{ashgrey}n/a & \color{ashgrey}52.2 \\
HuaweiTSC & 1 & 50.1 \\
\color{ashgrey}Online-Y & \color{ashgrey}n/a & \color{ashgrey}49.8 \\
\color{ashgrey}PROMT & \color{ashgrey}n/a & \color{ashgrey}49.6 \\
\color{ashgrey}ARC-NKUA & \color{ashgrey}n/a & \color{ashgrey}49.6 \\
ALMAnaCH-Inria & 2 & 21.8 \\
\bottomrule
\end{tabular}
\end{subtable}
\caption{Scores for the uk$\rightarrow$en translation task: chrF (nrefs:1|case:mixed|eff:yes|nc:6|nw:0|space:no|version:2.1.0), BLEU (nrefs:1|case:mixed|eff:no|tok:13a|smooth:exp|version:2.1.0), COMET (wmt20-comet-da).}
\end{table}
\begin{table} \definecolor{ashgrey}{rgb}{0.7, 0.75, 0.71}
\scriptsize\begin{subtable}[t]{0.33\textwidth}
\begin{tabular}{rccc}
\toprule
System & Rank & chrF \\
\midrule
\color{ashgrey}Online-B & \color{ashgrey}n/a & \color{ashgrey}59.3 \\
\color{ashgrey}Lan-Bridge & \color{ashgrey}n/a & \color{ashgrey}58.8 \\
\color{ashgrey}Online-G & \color{ashgrey}n/a & \color{ashgrey}57.2 \\
\color{ashgrey}Online-A & \color{ashgrey}n/a & \color{ashgrey}56.0 \\
\color{ashgrey}Online-Y & \color{ashgrey}n/a & \color{ashgrey}54.9 \\
eTranslation & 1 & 54.8 \\
HuaweiTSC & 2 & 54.8 \\
\color{ashgrey}ARC-NKUA & \color{ashgrey}n/a & \color{ashgrey}54.0 \\
\bottomrule
\end{tabular}
\end{subtable}
\begin{subtable}[t]{0.33\textwidth}
\begin{tabular}{rccc}
\toprule
System & Rank & BLEU \\
\midrule
\color{ashgrey}Online-B & \color{ashgrey}n/a & \color{ashgrey}32.5 \\
\color{ashgrey}Lan-Bridge & \color{ashgrey}n/a & \color{ashgrey}29.5 \\
\color{ashgrey}Online-A & \color{ashgrey}n/a & \color{ashgrey}28.0 \\
\color{ashgrey}Online-G & \color{ashgrey}n/a & \color{ashgrey}27.2 \\
\color{ashgrey}Online-Y & \color{ashgrey}n/a & \color{ashgrey}26.9 \\
HuaweiTSC & 1 & 26.5 \\
eTranslation & 2 & 26.2 \\
\color{ashgrey}ARC-NKUA & \color{ashgrey}n/a & \color{ashgrey}25.2 \\
\bottomrule
\end{tabular}
\end{subtable}
\begin{subtable}[t]{0.33\textwidth}
\begin{tabular}{rccc}
\toprule
System & Rank & COMET \\
\midrule
\color{ashgrey}Online-B & \color{ashgrey}n/a & \color{ashgrey}73.3 \\
\color{ashgrey}Online-G & \color{ashgrey}n/a & \color{ashgrey}69.9 \\
\color{ashgrey}Lan-Bridge & \color{ashgrey}n/a & \color{ashgrey}65.8 \\
\color{ashgrey}Online-A & \color{ashgrey}n/a & \color{ashgrey}60.9 \\
eTranslation & 1 & 54.6 \\
HuaweiTSC & 2 & 54.4 \\
\color{ashgrey}Online-Y & \color{ashgrey}n/a & \color{ashgrey}52.0 \\
\color{ashgrey}ARC-NKUA & \color{ashgrey}n/a & \color{ashgrey}49.2 \\
\bottomrule
\end{tabular}
\end{subtable}
\caption{Scores for the en$\rightarrow$uk translation task: chrF (nrefs:1|case:mixed|eff:yes|nc:6|nw:0|space:no|version:2.1.0), BLEU (nrefs:1|case:mixed|eff:no|tok:13a|smooth:exp|version:2.1.0), COMET (wmt20-comet-da).}
\end{table}

\begin{table} \definecolor{ashgrey}{rgb}{0.7, 0.75, 0.71}
\scriptsize\begin{subtable}[t]{0.33\textwidth}
\begin{tabular}{rccc}
\toprule
System & Rank & chrF \\
\midrule
JDExploreAcademy & 1 & 62.4 \\
\color{ashgrey}LanguageX & \color{ashgrey}n/a & \color{ashgrey}61.7 \\
\color{ashgrey}Online-G & \color{ashgrey}n/a & \color{ashgrey}60.9 \\
AISP-SJTU & 2 & 60.5 \\
\color{ashgrey}HuaweiTSC & \color{ashgrey}n/a & \color{ashgrey}59.6 \\
\color{ashgrey}Online-B & \color{ashgrey}n/a & \color{ashgrey}59.4 \\
\color{ashgrey}Lan-Bridge & \color{ashgrey}n/a & \color{ashgrey}59.1 \\
\color{ashgrey}Online-Y & \color{ashgrey}n/a & \color{ashgrey}58.6 \\
\color{ashgrey}Online-A & \color{ashgrey}n/a & \color{ashgrey}58.4 \\
NiuTrans & 3 & 57.2 \\
DLUT & 4 & 56.3 \\
\color{ashgrey}Online-W & \color{ashgrey}n/a & \color{ashgrey}55.6 \\
\bottomrule
\end{tabular}
\end{subtable}
\begin{subtable}[t]{0.33\textwidth}
\begin{tabular}{rccc}
\toprule
System & Rank & BLEU \\
\midrule
JDExploreAcademy & 1 & 37.9 \\
\color{ashgrey}LanguageX & \color{ashgrey}n/a & \color{ashgrey}36.4 \\
AISP-SJTU & 2 & 34.4 \\
\color{ashgrey}HuaweiTSC & \color{ashgrey}n/a & \color{ashgrey}34.3 \\
\color{ashgrey}Online-G & \color{ashgrey}n/a & \color{ashgrey}34.1 \\
\color{ashgrey}Online-B & \color{ashgrey}n/a & \color{ashgrey}33.3 \\
\color{ashgrey}Lan-Bridge & \color{ashgrey}n/a & \color{ashgrey}32.9 \\
\color{ashgrey}Online-A & \color{ashgrey}n/a & \color{ashgrey}31.6 \\
\color{ashgrey}Online-Y & \color{ashgrey}n/a & \color{ashgrey}31.1 \\
NiuTrans & 3 & 30.5 \\
DLUT & 4 & 29.0 \\
\color{ashgrey}Online-W & \color{ashgrey}n/a & \color{ashgrey}28.1 \\
\bottomrule
\end{tabular}
\end{subtable}
\begin{subtable}[t]{0.33\textwidth}
\begin{tabular}{rccc}
\toprule
System & Rank & COMET \\
\midrule
\color{ashgrey}Online-G & \color{ashgrey}n/a & \color{ashgrey}45.7 \\
JDExploreAcademy & 1 & 45.2 \\
\color{ashgrey}LanguageX & \color{ashgrey}n/a & \color{ashgrey}45.0 \\
\color{ashgrey}Lan-Bridge & \color{ashgrey}n/a & \color{ashgrey}43.1 \\
\color{ashgrey}HuaweiTSC & \color{ashgrey}n/a & \color{ashgrey}42.9 \\
\color{ashgrey}Online-B & \color{ashgrey}n/a & \color{ashgrey}42.2 \\
AISP-SJTU & 2 & 41.7 \\
\color{ashgrey}Online-Y & \color{ashgrey}n/a & \color{ashgrey}40.9 \\
\color{ashgrey}Online-A & \color{ashgrey}n/a & \color{ashgrey}35.3 \\
\color{ashgrey}Online-W & \color{ashgrey}n/a & \color{ashgrey}31.7 \\
NiuTrans & 3 & 31.4 \\
DLUT & 4 & 30.7 \\
\bottomrule
\end{tabular}
\end{subtable}
\caption{Scores for the zh$\rightarrow$en translation task: chrF (nrefs:2|case:mixed|eff:yes|nc:6|nw:0|space:no|version:2.1.0), BLEU (nrefs:2|case:mixed|eff:no|tok:13a|smooth:exp|version:2.1.0), COMET (wmt20-comet-da).}
\end{table}
\begin{table} \definecolor{ashgrey}{rgb}{0.7, 0.75, 0.71}
\scriptsize\begin{subtable}[t]{0.33\textwidth}
\begin{tabular}{rccc}
\toprule
System & Rank & chrF \\
\midrule
\color{ashgrey}Online-B & \color{ashgrey}n/a & \color{ashgrey}70.3 \\
\color{ashgrey}HuaweiTSC & \color{ashgrey}n/a & \color{ashgrey}61.0 \\
\color{ashgrey}LanguageX & \color{ashgrey}n/a & \color{ashgrey}59.5 \\
\color{ashgrey}Online-A & \color{ashgrey}n/a & \color{ashgrey}58.1 \\
Manifold & 1 & 57.7 \\
\color{ashgrey}Online-Y & \color{ashgrey}n/a & \color{ashgrey}57.0 \\
JDExploreAcademy & 2 & 55.8 \\
AISP-SJTU & 3 & 55.4 \\
\color{ashgrey}Lan-Bridge & \color{ashgrey}n/a & \color{ashgrey}53.7 \\
DLUT & 4 & 53.1 \\
\color{ashgrey}Online-G & \color{ashgrey}n/a & \color{ashgrey}52.4 \\
\color{ashgrey}Online-W & \color{ashgrey}n/a & \color{ashgrey}51.4 \\
\bottomrule
\end{tabular}
\end{subtable}
\begin{subtable}[t]{0.33\textwidth}
\begin{tabular}{rccc}
\toprule
System & Rank & BLEU \\
\midrule
\color{ashgrey}Online-B & \color{ashgrey}n/a & \color{ashgrey}80.0 \\
\color{ashgrey}HuaweiTSC & \color{ashgrey}n/a & \color{ashgrey}73.3 \\
\color{ashgrey}LanguageX & \color{ashgrey}n/a & \color{ashgrey}72.9 \\
Manifold & 1 & 69.4 \\
\color{ashgrey}Online-A & \color{ashgrey}n/a & \color{ashgrey}68.6 \\
JDExploreAcademy & 2 & 68.4 \\
\color{ashgrey}Online-Y & \color{ashgrey}n/a & \color{ashgrey}68.3 \\
AISP-SJTU & 3 & 67.6 \\
\color{ashgrey}Lan-Bridge & \color{ashgrey}n/a & \color{ashgrey}66.5 \\
DLUT & 4 & 63.9 \\
\color{ashgrey}Online-G & \color{ashgrey}n/a & \color{ashgrey}63.2 \\
\color{ashgrey}Online-W & \color{ashgrey}n/a & \color{ashgrey}62.0 \\
\bottomrule
\end{tabular}
\end{subtable}
\begin{subtable}[t]{0.33\textwidth}
\begin{tabular}{rccc}
\toprule
System & Rank & COMET \\
\midrule
\color{ashgrey}LanguageX & \color{ashgrey}n/a & \color{ashgrey}63.9 \\
\color{ashgrey}Online-B & \color{ashgrey}n/a & \color{ashgrey}61.8 \\
JDExploreAcademy & 1 & 61.8 \\
\color{ashgrey}Lan-Bridge & \color{ashgrey}n/a & \color{ashgrey}61.4 \\
\color{ashgrey}Online-W & \color{ashgrey}n/a & \color{ashgrey}61.0 \\
Manifold & 2 & 60.2 \\
\color{ashgrey}Online-Y & \color{ashgrey}n/a & \color{ashgrey}59.8 \\
\color{ashgrey}HuaweiTSC & \color{ashgrey}n/a & \color{ashgrey}59.6 \\
\color{ashgrey}Online-A & \color{ashgrey}n/a & \color{ashgrey}57.4 \\
AISP-SJTU & 3 & 56.6 \\
DLUT & 4 & 52.2 \\
\color{ashgrey}Online-G & \color{ashgrey}n/a & \color{ashgrey}51.3 \\
\bottomrule
\end{tabular}
\end{subtable}
\caption{\label{tab:lastranking} Scores for the en$\rightarrow$zh translation task: chrF (nrefs:2|case:mixed|eff:yes|nc:6|nw:0|space:no|version:2.1.0), BLEU (nrefs:2|case:mixed|eff:no|tok:zh|smooth:exp|version:2.1.0), COMET (wmt20-comet-da).}
\end{table}
\clearpage
\section{Statistical Significance Testing Tables}
From Table \ref{tab:firstsig} to \ref{tab:lastsig}.
\label{sec:sigtab}
\begin{table}
\scriptsize \begin{center}\begin{tabular}{rccccccccccc}
\toprule
& Onl. & JDE. & Lan. & Onl. & Onl. & CUN. & CUN. & Onl. & SHO. & Onl. & ALM. \\
\midrule
Online-W & \cellcolor{red!0} 0.0 & \cellcolor{red!70} \underline{4.7} & \cellcolor{red!70} \underline{5.1} & \cellcolor{red!70} \underline{5.2} & \cellcolor{red!70} \underline{5.6} & \cellcolor{red!70} \underline{6.7} & \cellcolor{red!70} \underline{7.3} & \cellcolor{red!70} \underline{8.6} & \cellcolor{red!70} \underline{9.6} & \cellcolor{red!70} \underline{10.9} & \cellcolor{red!70} \underline{21.4} \\
JDExploreAcademy &  & \cellcolor{red!0} 0.0 & \cellcolor{red!10} 0.4 & \cellcolor{red!30} \underline{0.5} & \cellcolor{red!70} \underline{1.0} & \cellcolor{red!70} \underline{2.1} & \cellcolor{red!70} \underline{2.7} & \cellcolor{red!70} \underline{4.0} & \cellcolor{red!70} \underline{4.9} & \cellcolor{red!70} \underline{6.3} & \cellcolor{red!70} \underline{16.7} \\
Lan-Bridge &  &  & \cellcolor{red!0} 0.0 & \cellcolor{red!0} 0.1 & \cellcolor{red!60} \underline{0.6} & \cellcolor{red!70} \underline{1.7} & \cellcolor{red!70} \underline{2.3} & \cellcolor{red!70} \underline{3.6} & \cellcolor{red!70} \underline{4.5} & \cellcolor{red!70} \underline{5.9} & \cellcolor{red!70} \underline{16.3} \\
Online-B &  &  &  & \cellcolor{red!0} 0.0 & \cellcolor{red!40} \underline{0.5} & \cellcolor{red!70} \underline{1.6} & \cellcolor{red!70} \underline{2.2} & \cellcolor{red!70} \underline{3.5} & \cellcolor{red!70} \underline{4.4} & \cellcolor{red!70} \underline{5.8} & \cellcolor{red!70} \underline{16.2} \\
Online-A &  &  &  &  & \cellcolor{red!0} 0.0 & \cellcolor{red!70} \underline{1.1} & \cellcolor{red!70} \underline{1.7} & \cellcolor{red!70} \underline{3.0} & \cellcolor{red!70} \underline{4.0} & \cellcolor{red!70} \underline{5.3} & \cellcolor{red!70} \underline{15.7} \\
\bottomrule
\end{tabular}
\caption{\label{tab:firstsig} Statistical significance testing of the chrF score difference for each system pair for the cs$\rightarrow$en.}
\end{center} \end{table}

\begin{table}
\scriptsize \begin{center}\begin{tabular}{rccccccccccc}
\toprule
& Onl. & JDE. & Lan. & Onl. & Onl. & CUN. & CUN. & Onl. & SHO. & Onl. & ALM. \\
\midrule
Online-W & \cellcolor{red!0} 0.0 & \cellcolor{red!70} \underline{7.8} & \cellcolor{red!70} \underline{7.9} & \cellcolor{red!70} \underline{8.1} & \cellcolor{red!70} \underline{9.2} & \cellcolor{red!70} \underline{10.8} & \cellcolor{red!70} \underline{11.1} & \cellcolor{red!70} \underline{14.4} & \cellcolor{red!70} \underline{15.2} & \cellcolor{red!70} \underline{17.7} & \cellcolor{red!70} \underline{31.8} \\
JDExploreAcademy &  & \cellcolor{red!0} 0.0 & \cellcolor{red!0} 0.1 & \cellcolor{red!0} 0.3 & \cellcolor{red!70} \underline{1.5} & \cellcolor{red!70} \underline{3.0} & \cellcolor{red!70} \underline{3.3} & \cellcolor{red!70} \underline{6.6} & \cellcolor{red!70} \underline{7.4} & \cellcolor{red!70} \underline{9.9} & \cellcolor{red!70} \underline{24.1} \\
Lan-Bridge &  &  & \cellcolor{red!0} 0.0 & \cellcolor{red!0} 0.2 & \cellcolor{red!60} \underline{1.3} & \cellcolor{red!70} \underline{2.9} & \cellcolor{red!70} \underline{3.2} & \cellcolor{red!70} \underline{6.5} & \cellcolor{red!70} \underline{7.3} & \cellcolor{red!70} \underline{9.8} & \cellcolor{red!70} \underline{23.9} \\
Online-B &  &  &  & \cellcolor{red!0} 0.0 & \cellcolor{red!60} \underline{1.2} & \cellcolor{red!70} \underline{2.7} & \cellcolor{red!70} \underline{3.0} & \cellcolor{red!70} \underline{6.3} & \cellcolor{red!70} \underline{7.1} & \cellcolor{red!70} \underline{9.6} & \cellcolor{red!70} \underline{23.8} \\
Online-A &  &  &  &  & \cellcolor{red!0} 0.0 & \cellcolor{red!60} \underline{1.5} & \cellcolor{red!70} \underline{1.8} & \cellcolor{red!70} \underline{5.1} & \cellcolor{red!70} \underline{5.9} & \cellcolor{red!70} \underline{8.4} & \cellcolor{red!70} \underline{22.6} \\
\bottomrule
\end{tabular}
\caption{Statistical significance testing of the BLEU score difference for each system pair for the cs$\rightarrow$en.}
\end{center} \end{table}

\begin{table}
\scriptsize \begin{center}\begin{tabular}{rccccccccccc}
\toprule
& Onl. & JDE. & Onl. & Lan. & CUN. & Onl. & CUN. & Onl. & SHO. & Onl. & ALM. \\
\midrule
Online-W & \cellcolor{red!0} 0.0 & \cellcolor{red!70} \underline{2.7} & \cellcolor{red!70} \underline{5.6} & \cellcolor{red!70} \underline{5.7} & \cellcolor{red!70} \underline{6.9} & \cellcolor{red!70} \underline{7.7} & \cellcolor{red!70} \underline{8.3} & \cellcolor{red!70} \underline{14.5} & \cellcolor{red!70} \underline{16.3} & \cellcolor{red!70} \underline{18.8} & \cellcolor{red!70} \underline{58.2}\\
JDExploreAcademy &  & \cellcolor{red!0} 0.0 & \cellcolor{red!70} \underline{2.9} & \cellcolor{red!70} \underline{3.0} & \cellcolor{red!70} \underline{4.2} & \cellcolor{red!70} \underline{5.0} & \cellcolor{red!70} \underline{5.6} & \cellcolor{red!70} \underline{11.8} & \cellcolor{red!70} \underline{13.6} & \cellcolor{red!70} \underline{16.1} & \cellcolor{red!70} \underline{55.5}\\
Online-B &  &  & \cellcolor{red!0} 0.0 & \cellcolor{red!0} 0.1 & \cellcolor{red!0} 1.3 & \cellcolor{red!70} \underline{2.1} & \cellcolor{red!70} \underline{2.7} & \cellcolor{red!70} \underline{8.9} & \cellcolor{red!70} \underline{10.7} & \cellcolor{red!70} \underline{13.2} & \cellcolor{red!70} \underline{52.6}\\
Lan-Bridge &  &  &  & \cellcolor{red!0} 0.0 & \cellcolor{red!0} 1.2 & \cellcolor{red!70} \underline{2.0} & \cellcolor{red!70} \underline{2.6} & \cellcolor{red!70} \underline{8.8} & \cellcolor{red!70} \underline{10.6} & \cellcolor{red!70} \underline{13.1} & \cellcolor{red!70} \underline{52.5}\\
CUNI-DocTransformer &  &  &  &  & \cellcolor{red!0} 0.0 & \cellcolor{red!0} 0.8 & \cellcolor{red!40} \underline{1.4} & \cellcolor{red!70} \underline{7.6} & \cellcolor{red!70} \underline{9.4} & \cellcolor{red!70} \underline{11.9} & \cellcolor{red!70} \underline{51.3}\\
\bottomrule
\end{tabular}
\caption{Statistical significance testing of the COMET score difference for each system pair for the cs$\rightarrow$en.}
 \end{center} \end{table}
 
 \begin{table}
\scriptsize \begin{center}\begin{tabular}{rcccccccccc}
\toprule
& Onl. & Lan. & Onl. & Onl. & JDE. & CUN. & CUN. & Onl. & Onl. & CUN. \\
\midrule
Online-B & \cellcolor{red!0} 0.0 & \cellcolor{red!70} \underline{0.9} & \cellcolor{red!50} \underline{0.9} & \cellcolor{red!70} \underline{3.8} & \cellcolor{red!70} \underline{4.1} & \cellcolor{red!70} \underline{5.3} & \cellcolor{red!70} \underline{6.2} & \cellcolor{red!70} \underline{6.7} & \cellcolor{red!70} \underline{6.8} & \cellcolor{red!70} \underline{7.1} \\
Lan-Bridge &  & \cellcolor{red!0} 0.0 & \cellcolor{red!0} -0.0 & \cellcolor{red!70} \underline{2.9} & \cellcolor{red!70} \underline{3.2} & \cellcolor{red!70} \underline{4.4} & \cellcolor{red!70} \underline{5.3} & \cellcolor{red!70} \underline{5.8} & \cellcolor{red!70} \underline{6.0} & \cellcolor{red!70} \underline{6.2} \\
Online-W &  &  & \cellcolor{red!0} 0.0 & \cellcolor{red!70} \underline{2.9} & \cellcolor{red!70} \underline{3.2} & \cellcolor{red!70} \underline{4.4} & \cellcolor{red!70} \underline{5.4} & \cellcolor{red!70} \underline{5.8} & \cellcolor{red!70} \underline{6.0} & \cellcolor{red!70} \underline{6.2} \\
Online-A &  &  &  & \cellcolor{red!0} 0.0 & \cellcolor{red!0} 0.3 & \cellcolor{red!70} \underline{1.6} & \cellcolor{red!70} \underline{2.5} & \cellcolor{red!70} \underline{2.9} & \cellcolor{red!70} \underline{3.1} & \cellcolor{red!70} \underline{3.3} \\
JDExploreAcademy &  &  &  &  & \cellcolor{red!0} 0.0 & \cellcolor{red!70} \underline{1.2} & \cellcolor{red!70} \underline{2.1} & \cellcolor{red!70} \underline{2.6} & \cellcolor{red!70} \underline{2.8} & \cellcolor{red!70} \underline{3.0} \\
CUNI-DocTransformer &  &  &  &  &  & \cellcolor{red!0} 0.0 & \cellcolor{red!70} \underline{0.9} & \cellcolor{red!70} \underline{1.4} & \cellcolor{red!70} \underline{1.5} & \cellcolor{red!70} \underline{1.8} \\
CUNI-Bergamot &  &  &  &  &  &  & \cellcolor{red!0} 0.0 & \cellcolor{red!30} \underline{0.5} & \cellcolor{red!50} \underline{0.6} & \cellcolor{red!70} \underline{0.9} \\
\bottomrule
\end{tabular}
\caption{Statistical significance testing of the chrF score difference for each system pair for the en$\rightarrow$cs.}
\end{center} \end{table}

\begin{table}
\scriptsize \begin{center}\begin{tabular}{rcccccccccc}
\toprule
& Onl. & Lan. & Onl. & JDE. & Onl. & CUN. & CUN. & CUN. & Onl. & Onl. \\
\midrule
Online-B & \cellcolor{red!0} 0.0 & \cellcolor{red!70} \underline{3.1} & \cellcolor{red!70} \underline{3.4} & \cellcolor{red!70} \underline{7.0} & \cellcolor{red!70} \underline{7.4} & \cellcolor{red!70} \underline{8.7} & \cellcolor{red!70} \underline{10.2} & \cellcolor{red!70} \underline{10.9} & \cellcolor{red!70} \underline{11.6} & \cellcolor{red!70} \underline{11.8} \\
Lan-Bridge &  & \cellcolor{red!0} 0.0 & \cellcolor{red!0} 0.3 & \cellcolor{red!70} \underline{3.9} & \cellcolor{red!70} \underline{4.3} & \cellcolor{red!70} \underline{5.6} & \cellcolor{red!70} \underline{7.1} & \cellcolor{red!70} \underline{7.8} & \cellcolor{red!70} \underline{8.5} & \cellcolor{red!70} \underline{8.7} \\
Online-W &  &  & \cellcolor{red!0} 0.0 & \cellcolor{red!70} \underline{3.7} & \cellcolor{red!70} \underline{4.0} & \cellcolor{red!70} \underline{5.3} & \cellcolor{red!70} \underline{6.9} & \cellcolor{red!70} \underline{7.5} & \cellcolor{red!70} \underline{8.3} & \cellcolor{red!70} \underline{8.4} \\
JDExploreAcademy &  &  &  & \cellcolor{red!0} 0.0 & \cellcolor{red!0} 0.3 & \cellcolor{red!70} \underline{1.6} & \cellcolor{red!70} \underline{3.2} & \cellcolor{red!70} \underline{3.9} & \cellcolor{red!70} \underline{4.6} & \cellcolor{red!70} \underline{4.7} \\
Online-A &  &  &  &  & \cellcolor{red!0} 0.0 & \cellcolor{red!60} \underline{1.3} & \cellcolor{red!70} \underline{2.9} & \cellcolor{red!70} \underline{3.5} & \cellcolor{red!70} \underline{4.3} & \cellcolor{red!70} \underline{4.4} \\
CUNI-DocTransformer &  &  &  &  &  & \cellcolor{red!0} 0.0 & \cellcolor{red!70} \underline{1.6} & \cellcolor{red!70} \underline{2.2} & \cellcolor{red!70} \underline{3.0} & \cellcolor{red!70} \underline{3.1} \\
CUNI-Bergamot &  &  &  &  &  &  & \cellcolor{red!0} 0.0 & \cellcolor{red!20} \underline{0.7} & \cellcolor{red!70} \underline{1.4} & \cellcolor{red!70} \underline{1.5} \\
\bottomrule
\end{tabular}
\caption{Statistical significance testing of the BLEU score difference for each system pair for the en$\rightarrow$cs.}
\end{center} \end{table}

\begin{table}
\scriptsize \begin{center}\begin{tabular}{rcccccccccc}
\toprule
& Onl. & Onl. & CUN. & JDE. & Lan. & Onl. & CUN. & CUN. & Onl. & Onl. \\
\midrule
Online-W & \cellcolor{red!0} 0.0 & \cellcolor{red!0} 0.4 & \cellcolor{red!60} \underline{1.8} & \cellcolor{red!70} \underline{2.6} & \cellcolor{red!70} \underline{3.2} & \cellcolor{red!70} \underline{5.7} & \cellcolor{red!70} \underline{6.2} & \cellcolor{red!70} \underline{11.2} & \cellcolor{red!70} \underline{14.2} & \cellcolor{red!70} \underline{15.6}\\
Online-B &  & \cellcolor{red!0} 0.0 & \cellcolor{red!40} \underline{1.4} & \cellcolor{red!70} \underline{2.2} & \cellcolor{red!70} \underline{2.8} & \cellcolor{red!70} \underline{5.3} & \cellcolor{red!70} \underline{5.8} & \cellcolor{red!70} \underline{10.8} & \cellcolor{red!70} \underline{13.8} & \cellcolor{red!70} \underline{15.2}\\
CUNI-Bergamot &  &  & \cellcolor{red!0} 0.0 & \cellcolor{red!0} 0.8 & \cellcolor{red!20} \underline{1.4} & \cellcolor{red!70} \underline{3.9} & \cellcolor{red!70} \underline{4.4} & \cellcolor{red!70} \underline{9.4} & \cellcolor{red!70} \underline{12.4} & \cellcolor{red!70} \underline{13.8}\\
JDExploreAcademy &  &  &  & \cellcolor{red!0} 0.0 & \cellcolor{red!0} 0.6 & \cellcolor{red!70} \underline{3.1} & \cellcolor{red!70} \underline{3.6} & \cellcolor{red!70} \underline{8.6} & \cellcolor{red!70} \underline{11.6} & \cellcolor{red!70} \underline{13.0}\\
Lan-Bridge &  &  &  &  & \cellcolor{red!0} 0.0 & \cellcolor{red!70} \underline{2.5} & \cellcolor{red!70} \underline{3.0} & \cellcolor{red!70} \underline{8.0} & \cellcolor{red!70} \underline{11.0} & \cellcolor{red!70} \underline{12.4}\\
Online-A &  &  &  &  &  & \cellcolor{red!0} 0.0 & \cellcolor{red!0} 0.5 & \cellcolor{red!70} \underline{5.5} & \cellcolor{red!70} \underline{8.5} & \cellcolor{red!70} \underline{9.9}\\
CUNI-DocTransformer &  &  &  &  &  &  & \cellcolor{red!0} 0.0 & \cellcolor{red!70} \underline{5.0} & \cellcolor{red!70} \underline{8.0} & \cellcolor{red!70} \underline{9.4}\\
\bottomrule
\end{tabular}
\caption{Statistical significance testing of the COMET score difference for each system pair for the en$\rightarrow$cs.}
 \end{center} \end{table}

\begin{table}
\scriptsize \begin{center}%
\caption{Statistical significance testing of the chrF score difference for each system pair for the cs$\rightarrow$uk.}
\end{center} \end{table}

\begin{table}
\scriptsize \begin{center}%
\caption{Statistical significance testing of the BLEU score difference for each system pair for the cs$\rightarrow$uk.}
\end{center} \end{table}

\begin{table}
\scriptsize \begin{center}%
\caption{Statistical significance testing of the COMET score difference for each system pair for the cs$\rightarrow$uk.}
 \end{center} \end{table}
 \begin{table}
\scriptsize \begin{center}%
\caption{Statistical significance testing of the chrF score difference for each system pair for the uk$\rightarrow$cs.}
\end{center} \end{table}

\begin{table}
\scriptsize \begin{center}%
\caption{Statistical significance testing of the BLEU score difference for each system pair for the uk$\rightarrow$cs.}
\end{center} \end{table}

\begin{table}
\scriptsize \begin{center}%
\caption{Statistical significance testing of the COMET score difference for each system pair for the uk$\rightarrow$cs.}
 \end{center} \end{table}
 
 \clearpage
 \begin{table}
\scriptsize \begin{center}%
\caption{Statistical significance testing of the chrF score difference for each system pair for the de$\rightarrow$en.}
\end{center} \end{table}

\begin{table}
\scriptsize \begin{center}%
\caption{Statistical significance testing of the BLEU score difference for each system pair for the de$\rightarrow$en.}
\end{center} \end{table}

\begin{table}
\scriptsize \begin{center}%
\caption{Statistical significance testing of the COMET score difference for each system pair for the de$\rightarrow$en.}
 \end{center} \end{table}
 \begin{table}
\scriptsize \begin{center}%
\caption{Statistical significance testing of the chrF score difference for each system pair for the en$\rightarrow$de.}
\end{center} \end{table}

\begin{table}
\scriptsize \begin{center}%
\caption{Statistical significance testing of the BLEU score difference for each system pair for the en$\rightarrow$de.}
\end{center} \end{table}

\begin{table}
\scriptsize \begin{center}%
\caption{Statistical significance testing of the COMET score difference for each system pair for the en$\rightarrow$de.}
 \end{center} \end{table}
 
 \begin{table}
\scriptsize \begin{center}%
\caption{Statistical significance testing of the chrF score difference for each system pair for the de$\rightarrow$fr.}
\end{center} \end{table}

\begin{table}
\scriptsize \begin{center}%
\caption{Statistical significance testing of the BLEU score difference for each system pair for the de$\rightarrow$fr.}
\end{center} \end{table}

\begin{table}
\scriptsize \begin{center}%
\caption{Statistical significance testing of the COMET score difference for each system pair for the de$\rightarrow$fr.}
 \end{center} \end{table}
 \begin{table}
\scriptsize \begin{center}%
\caption{Statistical significance testing of the chrF score difference for each system pair for the fr$\rightarrow$de.}
\end{center} \end{table}

\begin{table}
\scriptsize \begin{center}%
\caption{Statistical significance testing of the BLEU score difference for each system pair for the fr$\rightarrow$de.}
\end{center} \end{table}

\begin{table}
\scriptsize \begin{center}%
\caption{Statistical significance testing of the COMET score difference for each system pair for the fr$\rightarrow$de.}
 \end{center} \end{table}
 
 \begin{table}
\scriptsize \begin{center}%
\caption{Statistical significance testing of the chrF score difference for each system pair for the en$\rightarrow$hr.}
\end{center} \end{table}

\begin{table}
\scriptsize \begin{center}%
\caption{Statistical significance testing of the BLEU score difference for each system pair for the en$\rightarrow$hr.}
\end{center} \end{table}

\begin{table}
\scriptsize \begin{center}%
\caption{Statistical significance testing of the COMET score difference for each system pair for the en$\rightarrow$hr.}
 \end{center} \end{table}

 \begin{table}
\scriptsize \begin{center}%
\caption{Statistical significance testing of the chrF score difference for each system pair for the ja$\rightarrow$en.}
\end{center} \end{table}

\begin{table}
\scriptsize \begin{center}%
\caption{Statistical significance testing of the BLEU score difference for each system pair for the ja$\rightarrow$en.}
\end{center} \end{table}
\begin{table}
\scriptsize \begin{center}%
\caption{Statistical significance testing of the COMET score difference for each system pair for the ja$\rightarrow$en.}
 \end{center} \end{table}

 \begin{table}
\scriptsize \begin{center}%
\caption{Statistical significance testing of the chrF score difference for each system pair for the en$\rightarrow$ja.}
\end{center} \end{table}

\begin{table}
\scriptsize \begin{center}%
\caption{Statistical significance testing of the BLEU score difference for each system pair for the en$\rightarrow$ja.}
\end{center} \end{table}

\begin{table}
\scriptsize \begin{center}%
\caption{Statistical significance testing of the COMET score difference for each system pair for the en$\rightarrow$ja.}
 \end{center} \end{table}
 
 \begin{table}
\scriptsize \begin{center}%
\caption{Statistical significance testing of the chrF score difference for each system pair for the liv$\rightarrow$en.}
\end{center} \end{table}

\begin{table}
\scriptsize \begin{center}%
\caption{Statistical significance testing of the BLEU score difference for each system pair for the liv$\rightarrow$en.}
\end{center} \end{table}

\begin{table}
\scriptsize \begin{center}%
\caption{Statistical significance testing of the COMET score difference for each system pair for the liv$\rightarrow$en.}
 \end{center} \end{table}
 
 \begin{table}
\scriptsize \begin{center}%
\caption{Statistical significance testing of the chrF score difference for each system pair for the en$\rightarrow$liv.}
\end{center} \end{table}

\begin{table}
\scriptsize \begin{center}%
\caption{Statistical significance testing of the BLEU score difference for each system pair for the en$\rightarrow$liv.}
\end{center} \end{table}

\begin{table}
\scriptsize \begin{center}%
\caption{Statistical significance testing of the COMET score difference for each system pair for the en$\rightarrow$liv.}
 \end{center} \end{table}
 
 \begin{table}
\scriptsize \begin{center}%
\caption{Statistical significance testing of the chrF score difference for each system pair for the ru$\rightarrow$en.}
\end{center} \end{table}

\begin{table}
\scriptsize \begin{center}%
\caption{Statistical significance testing of the BLEU score difference for each system pair for the ru$\rightarrow$en.}
\end{center} \end{table}

\begin{table}
\scriptsize \begin{center}%
\caption{Statistical significance testing of the COMET score difference for each system pair for the ru$\rightarrow$en.}
 \end{center} \end{table}

 \begin{table}
\scriptsize \begin{center}%
\caption{Statistical significance testing of the chrF score difference for each system pair for the en$\rightarrow$ru.}
\end{center} \end{table}

\begin{table}
\scriptsize \begin{center}%
\caption{Statistical significance testing of the BLEU score difference for each system pair for the en$\rightarrow$ru.}
\end{center} \end{table}

\begin{table}
\scriptsize \begin{center}%
\caption{Statistical significance testing of the COMET score difference for each system pair for the en$\rightarrow$ru.}
 \end{center} \end{table}

 \begin{table}
\scriptsize \begin{center}\begin{tabular}{rcc}
\toprule
& Lan. & Onl. \\
\midrule
Lan-Bridge & \cellcolor{red!0} 0.0 & \cellcolor{red!70} \underline{9.7} \\
Online-G &  & \cellcolor{red!0} 0.0 \\
\bottomrule
\end{tabular}
\caption{Statistical significance testing of the chrF score difference for each system pair for the ru$\rightarrow$sah.}
\end{center} \end{table}

\begin{table}
\scriptsize \begin{center}%
\caption{Statistical significance testing of the BLEU score difference for each system pair for the ru$\rightarrow$sah.}
\end{center} \end{table}

\begin{table}
\scriptsize \begin{center}%
\caption{Statistical significance testing of the COMET score difference for each system pair for the ru$\rightarrow$sah.}
 \end{center} \end{table}

 \begin{table}
\scriptsize \begin{center}%
\caption{Statistical significance testing of the chrF score difference for each system pair for the sah$\rightarrow$ru.}
\end{center} \end{table}

\begin{table}
\scriptsize \begin{center}%
\caption{Statistical significance testing of the BLEU score difference for each system pair for the sah$\rightarrow$ru.}
\end{center} \end{table}

\begin{table}
\scriptsize \begin{center}%
\caption{Statistical significance testing of the COMET score difference for each system pair for the sah$\rightarrow$ru.}
 \end{center} \end{table}

 \begin{table}
\scriptsize \begin{center}%
\caption{Statistical significance testing of the chrF score difference for each system pair for the uk$\rightarrow$en.}
\end{center} \end{table}

\begin{table}
\scriptsize \begin{center}%
\caption{Statistical significance testing of the BLEU score difference for each system pair for the uk$\rightarrow$en.}
\end{center} \end{table}

\begin{table}
\scriptsize \begin{center}%
\caption{Statistical significance testing of the COMET score difference for each system pair for the uk$\rightarrow$en.}
 \end{center} \end{table}
 
 \begin{table}
\scriptsize \begin{center}%
\caption{Statistical significance testing of the chrF score difference for each system pair for the en$\rightarrow$uk.}
\end{center} \end{table}

\begin{table}
\scriptsize \begin{center}%
\caption{Statistical significance testing of the BLEU score difference for each system pair for the en$\rightarrow$uk.}
\end{center} \end{table}

\begin{table}
\scriptsize \begin{center}%
\caption{Statistical significance testing of the COMET score difference for each system pair for the en$\rightarrow$uk.}
 \end{center} \end{table}
 
 \begin{table}
\scriptsize \begin{center}%
\caption{Statistical significance testing of the chrF score difference for each system pair for the zh$\rightarrow$en.}
\end{center} \end{table}

\begin{table}
\scriptsize \begin{center}%
\caption{Statistical significance testing of the BLEU score difference for each system pair for the zh$\rightarrow$en.}
\end{center} \end{table}

\begin{table}
\scriptsize \begin{center}%
\caption{Statistical significance testing of the COMET score difference for each system pair for the zh$\rightarrow$en.}
 \end{center} \end{table}

 \begin{table}
\scriptsize \begin{center}%
\caption{Statistical significance testing of the chrF score difference for each system pair for the en$\rightarrow$zh.}
\end{center} \end{table}

\begin{table}
\scriptsize \begin{center}%
\caption{Statistical significance testing of the BLEU score difference for each system pair for the en$\rightarrow$zh.}
\end{center} \end{table}

\begin{table}
\scriptsize \begin{center}%
\caption{\label{tab:lastsig} Statistical significance testing of the COMET score difference for each system pair for the en$\rightarrow$zh.}
 \end{center} \end{table}
 \clearpage
 \section{MBR and Oracle Combination Tables}
 \label{sec:mbrtab}
 From Table \ref{tab:firstmbr} to \ref{tab:lastmbr}.

 \begin{table}
\scriptsize 
\begin{subtable}[t]{0.49\textwidth}\begin{center}
%
 \end{center}
\caption{cs$\rightarrow$en}
\end{subtable}
\begin{subtable}[t]{0.49\textwidth}\begin{center}
%
 \end{center}
\caption{en$\rightarrow$cs}
\end{subtable}
\caption{\label{tab:firstmbr}MBR and oracle combination using COMET for cs-en.}
 \end{table}

  \begin{table}
\scriptsize 
\begin{subtable}[t]{0.49\textwidth}\begin{center}
%
 \end{center}
\caption{cs$\rightarrow$uk}
\end{subtable}
\begin{subtable}[t]{0.49\textwidth}\begin{center}
%
 \end{center}
\caption{uk$\rightarrow$cs}
\end{subtable}
\caption{MBR and oracle combination using COMET for cs-uk.}
 \end{table}

  \begin{table}
\scriptsize 
\begin{subtable}[t]{0.49\textwidth}\begin{center}
%
 \end{center}
\caption{de$\rightarrow$en}
\end{subtable}
\begin{subtable}[t]{0.49\textwidth}\begin{center}
%
 \end{center}
\caption{en$\rightarrow$de}
\end{subtable}
\caption{MBR and oracle combination using COMET for de-en.}
 \end{table}

  \begin{table}
\scriptsize 
\begin{subtable}[t]{0.49\textwidth}\begin{center}
%
 \end{center}
\caption{de$\rightarrow$fr}
\end{subtable}
\begin{subtable}[t]{0.49\textwidth}\begin{center}
%
 \end{center}
\caption{en$\rightarrow$zh}
\end{subtable}
\caption{\label{tab:lastmbr}MBR and oracle combination using COMET for en-zh.}
 \end{table}
 
  \clearpage
 \section{Rankings With Different COMET Models}
 \label{sec:comettab}
From Table \ref{tab:firstcomet} to \ref{tab:lastcomet}.
   \begin{table}
\scriptsize 
\begin{subtable}[t]{0.49\textwidth}
%
 \caption{cs$\rightarrow$en}
\end{subtable}

\begin{subtable}[t]{0.49\textwidth}
%
 \caption{en$\rightarrow$zh}
\end{subtable}

\caption{ \label{tab:lastcomet} COMET scores (ranking) with different models for en-zh.}
 \end{table}
 
\section{Sentence Matching Statistic Tables}
\label{sec:matchtab}
From Table \ref{tab:firstmatch} to \ref{tab:lastmatch}.
    \begin{table}
\scriptsize 
\begin{subtable}[t]{0.49\textwidth}
\scalebox{0.85}{%
} \caption{en$\rightarrow$cs}
\end{subtable}

\caption{ \label{tab:firstmatch} Sentence Matching Statistics. ``Exact Match'' counts how many segments in the translation are identical to the corresponding segments of the reference translation. ``Self Mismatch'' provides various statistics: the number of non-unique source segments translated differently by the system/the number of non-unique source segments translated differently by the reference/the number of duplicated source segments/the total number of source segments. ``BLEU'' and ``COMET'' give the scores computed on the original/normalized translations.}
 \end{table}

     \begin{table}
\scriptsize 
\begin{subtable}[t]{0.49\textwidth}
\scalebox{0.85}{\begin{tabular}{rcccc}
\toprule
System & Exact Match & Self Mismatch & BLEU & COMET \\
\midrule
AMU & 24  & 0/0/0/1930 & 34.7/34.3  & 99.4/99.4 \\
Online-B & 80  & 0/0/0/1930 & 38.3/37.6  & 94.3/94.2 \\
Lan-Bridge & 66  & 0/0/0/1930 & 38.3/38.2  & 91.8/91.8 \\
CharlesTranslator & 46  & 0/0/0/1930 & 34.3/34.0  & 90.8/90.7 \\
HuaweiTSC & 51  & 0/0/0/1930 & 36.0/35.7  & 90.7/90.7 \\
CUNI-JL-JH & 39  & 0/0/0/1930 & 34.8/34.4  & 90.1/90.0 \\
Online-G & 29  & 0/0/0/1930 & 32.5/32.5  & 88.4/88.3 \\
Online-A & 43  & 0/0/0/1930 & 35.9/35.5  & 87.9/87.8 \\
CUNI-Transformer & 42  & 0/0/0/1930 & 35.0/34.8  & 87.4/87.3 \\
Online-Y & 40  & 0/0/0/1930 & 32.1/31.9  & 78.4/78.4 \\
ALMAnaCH-Inria & 21  & 0/0/0/1930 & 26.8/26.7  & 61.4/61.3 \\
\bottomrule
\end{tabular}} \caption{cs$\rightarrow$uk}
\end{subtable}
\begin{subtable}[t]{0.49\textwidth}
\scalebox{0.85}{\begin{tabular}{rcccc}
\toprule
System & Exact Match & Self Mismatch & BLEU & COMET \\
\midrule
AMU & 117  & 4/0/24/2812 & 37.0/36.4  & 104.9/104.7 \\
Online-B & 127  & 0/0/24/2812 & 36.4/35.7  & 96.5/96.3 \\
Lan-Bridge & 147  & 0/0/24/2812 & 36.5/36.5  & 94.6/94.6 \\
HuaweiTSC & 132  & 0/0/24/2812 & 36.0/35.7  & 91.5/91.4 \\
CharlesTranslator & 125  & 0/0/24/2812 & 35.9/35.0  & 90.2/90.0 \\
CUNI-JL-JH & 122  & 0/0/24/2812 & 35.1/34.6  & 89.1/89.0 \\
CUNI-Transformer & 126  & 0/0/24/2812 & 35.8/34.8  & 88.5/88.4 \\
Online-A & 116  & 0/0/24/2812 & 33.3/33.0  & 85.4/85.3 \\
Online-G & 106  & 0/0/24/2812 & 31.5/31.4  & 84.2/84.2 \\
Online-Y & 88  & 0/0/24/2812 & 29.6/29.1  & 78.7/78.4 \\
ALMAnaCH-Inria & 75  & 0/0/24/2812 & 25.3/25.0  & 62.5/62.3 \\
\bottomrule
\end{tabular}} \caption{uk$\rightarrow$cs}
\end{subtable}

\caption{ Sentence Matching Statistics. ``Exact Match'' counts how many segments in the translation are identical to the corresponding segments of the reference translation. ``Self Mismatch'' provides various statistics: the number of non-unique source segments translated differently by the system/the number of non-unique source segments translated differently by the reference/the number of duplicated source segments/the total number of source segments. ``BLEU'' and ``COMET'' give the scores computed on the original/normalized translations.}
 \end{table}

      \begin{table}
\scriptsize 
\begin{subtable}[t]{0.49\textwidth}
\scalebox{0.85}{\begin{tabular}{rcccc}
\toprule
System & Exact Match & Self Mismatch & BLEU & COMET \\
\midrule
JDExploreAcademy & 62  & 0/1/5/1984 & 49.3/49.3  & 58.0/58.0 \\
Online-B & 58  & 0/1/5/1984 & 49.7/49.7  & 57.0/57.0 \\
Lan-Bridge & 67  & 0/1/5/1984 & 50.1/50.1  & 56.6/56.6 \\
Online-G & 59  & 0/1/5/1984 & 49.7/49.7  & 55.3/55.3 \\
Online-Y & 45  & 0/1/5/1984 & 49.3/48.6  & 54.7/54.6 \\
Online-A & 61  & 0/1/5/1984 & 50.2/50.2  & 54.5/54.5 \\
Online-W & 66  & 1/1/5/1984 & 48.8/48.8  & 54.4/54.4 \\
PROMT & 59  & 0/1/5/1984 & 49.2/49.2  & 51.8/51.8 \\
LT22 & 27  & 0/1/5/1984 & 40.3/40.3  & 25.6/25.6 \\
\bottomrule
\end{tabular}} \caption{de$\rightarrow$en}
\end{subtable}
\begin{subtable}[t]{0.49\textwidth}
\scalebox{0.85}{\begin{tabular}{rcccc}
\toprule
System & Exact Match & Self Mismatch & BLEU & COMET \\
\midrule
Online-W & 75  & 29/0/122/2037 & 48.9/48.8  & 65.6/65.6 \\
JDExploreAcademy & 70  & 0/0/122/2037 & 51.6/50.2  & 63.3/63.1 \\
Online-B & 74  & 0/0/122/2037 & 52.3/51.0  & 62.3/62.2 \\
Online-Y & 55  & 0/0/122/2037 & 50.5/49.1  & 61.1/60.9 \\
Online-A & 75  & 0/0/122/2037 & 50.1/50.1  & 60.6/60.6 \\
Online-G & 47  & 0/0/122/2037 & 49.6/49.6  & 60.2/60.2 \\
Lan-Bridge & 70  & 0/0/122/2037 & 49.4/49.4  & 58.8/58.8 \\
OpenNMT & 86  & 0/0/122/2037 & 48.4/46.7  & 57.3/56.9 \\
PROMT & 64  & 0/0/122/2037 & 49.0/47.7  & 55.9/55.7 \\
\bottomrule
\end{tabular}} \caption{en$\rightarrow$de}
\end{subtable}

\caption{ Sentence Matching Statistics. ``Exact Match'' counts how many segments in the translation are identical to the corresponding segments of the reference translation. ``Self Mismatch'' provides various statistics: the number of non-unique source segments translated differently by the system/the number of non-unique source segments translated differently by the reference/the number of duplicated source segments/the total number of source segments. ``BLEU'' and ``COMET'' give the scores computed on the original/normalized translations.}
 \end{table}

      \begin{table}
\scriptsize 
\begin{subtable}[t]{0.49\textwidth}
\scalebox{0.85}{\begin{tabular}{rcccc}
\toprule
System & Exact Match & Self Mismatch & BLEU & COMET \\
\midrule
Online-B & 264  & 0/0/5/1984 & 58.4/58.3  & 70.5/70.4 \\
Online-W & 90  & 3/0/5/1984 & 43.6/43.6  & 63.7/63.6 \\
Online-Y & 80  & 0/0/5/1984 & 46.2/45.2  & 57.8/57.3 \\
Online-A & 62  & 0/0/5/1984 & 41.3/45.2  & 52.3/52.5 \\
Online-G & 55  & 0/0/5/1984 & 39.0/39.0  & 44.9/44.9 \\
LT22 & 23  & 0/0/5/1984 & 28.3/28.5  & 10.5/10.5 \\
\bottomrule
\end{tabular}} \caption{de$\rightarrow$fr}
\end{subtable}
\begin{subtable}[t]{0.49\textwidth}
\scalebox{0.85}{\begin{tabular}{rcccc}
\toprule
System & Exact Match & Self Mismatch & BLEU & COMET \\
\midrule
Online-W & 315  & 8/0/31/2006 & 64.8/64.8  & 77.9/77.9 \\
Online-B & 153  & 0/0/31/2006 & 46.6/46.2  & 63.8/63.6 \\
Online-Y & 121  & 0/0/31/2006 & 45.0/44.3  & 61.6/61.4 \\
Online-A & 133  & 0/0/31/2006 & 44.4/44.4  & 59.3/59.3 \\
eTranslation & 162  & 0/0/31/2006 & 46.5/46.0  & 55.5/55.2 \\
Lan-Bridge & 107  & 0/0/31/2006 & 41.8/41.8  & 51.2/51.2 \\
Online-G & 85  & 0/0/31/2006 & 41.1/41.0  & 48.3/48.4 \\
\bottomrule
\end{tabular}} \caption{fr$\rightarrow$de}
\end{subtable}

\caption{ Sentence Matching Statistics. ``Exact Match'' counts how many segments in the translation are identical to the corresponding segments of the reference translation. ``Self Mismatch'' provides various statistics: the number of non-unique source segments translated differently by the system/the number of non-unique source segments translated differently by the reference/the number of duplicated source segments/the total number of source segments. ``BLEU'' and ``COMET'' give the scores computed on the original/normalized translations.}
 \end{table}

       \begin{table}
\scriptsize 
\begin{subtable}[t]{0.49\textwidth}
\scalebox{0.85}{\begin{tabular}{rcccc}
\toprule
System & Exact Match & Self Mismatch & BLEU & COMET \\
\midrule
Online-B & 80  & 0/5/11/1671 & 31.5/31.5  & 80.4/80.4 \\
Lan-Bridge & 77  & 0/5/11/1671 & 31.5/31.4  & 79.6/79.6 \\
Online-A & 64  & 0/5/11/1671 & 29.1/29.1  & 69.5/69.5 \\
SRPOL & 61  & 0/5/11/1671 & 29.1/29.1  & 69.5/69.5 \\
HuaweiTSC & 62  & 0/5/11/1671 & 29.9/29.9  & 67.7/67.7 \\
NiuTrans & 65  & 0/5/11/1671 & 29.3/29.3  & 65.6/65.6 \\
Online-G & 48  & 0/5/11/1671 & 25.7/25.7  & 64.3/64.3 \\
Online-Y & 54  & 0/5/11/1671 & 26.6/26.5  & 56.8/56.7 \\
\bottomrule
\end{tabular}} \caption{en$\rightarrow$hr}
\end{subtable}

\caption{ Sentence Matching Statistics. ``Exact Match'' counts how many segments in the translation are identical to the corresponding segments of the reference translation. ``Self Mismatch'' provides various statistics: the number of non-unique source segments translated differently by the system/the number of non-unique source segments translated differently by the reference/the number of duplicated source segments/the total number of source segments. ``BLEU'' and ``COMET'' give the scores computed on the original/normalized translations.}
 \end{table}

        \begin{table}
\scriptsize 
\begin{subtable}[t]{0.49\textwidth}
\scalebox{0.85}{\begin{tabular}{rcccc}
\toprule
System & Exact Match & Self Mismatch & BLEU & COMET \\
\midrule
NAIST-NICT-TIT & 35  & 13/11/53/2008 & 22.7/22.9  & 33.4/33.4 \\
AIST & 0  & 24/11/53/2008 & 0.1/0.1  & -152.7/-152.6 \\
NT5 & 48  & 0/11/53/2008 & 26.6/26.6  & 42.1/42.1 \\
Online-W & 45  & 14/11/53/2008 & 27.8/27.7  & 41.2/41.2 \\
JDExploreAcademy & 75  & 0/11/53/2008 & 25.6/25.6  & 40.6/40.6 \\
Online-B & 63  & 0/11/53/2008 & 24.7/24.8  & 39.7/39.6 \\
DLUT & 66  & 0/11/53/2008 & 24.8/24.8  & 37.3/37.3 \\
Online-A & 39  & 0/11/53/2008 & 22.8/22.9  & 32.9/32.9 \\
LanguageX & 36  & 0/11/53/2008 & 22.4/22.3  & 33.0/32.9 \\
Online-Y & 32  & 0/11/53/2008 & 21.5/22.4  & 32.4/32.3 \\
Lan-Bridge & 44  & 0/11/53/2008 & 22.8/22.7  & 32.0/31.9 \\
AISP-SJTU & 36  & 0/11/53/2008 & 22.0/21.9  & 30.2/30.2 \\
Online-G & 34  & 0/11/53/2008 & 19.7/20.1  & 22.3/22.5 \\
KYB & 30  & 0/11/53/2008 & 18.1/18.2  & 17.4/17.5 \\
\bottomrule
\end{tabular}} \caption{ja$\rightarrow$en}
\end{subtable}
\begin{subtable}[t]{0.49\textwidth}
\scalebox{0.85}{\begin{tabular}{rcccc}
\toprule
System & Exact Match & Self Mismatch & BLEU & COMET \\
\midrule
JDExploreAcademy & 10  & 0/63/122/2037 & 25.4/2.9  & 65.2/59.9 \\
NT5 & 15  & 0/63/122/2037 & 18.2/2.9  & 64.1/59.1 \\
LanguageX & 15  & 1/63/122/2037 & 24.5/2.7  & 62.1/57.1 \\
Online-B & 12  & 0/63/122/2037 & 8.0/2.4  & 60.8/57.1 \\
DLUT & 15  & 0/63/122/2037 & 12.8/2.7  & 60.5/55.3 \\
Online-W & 14  & 47/63/122/2037 & 17.6/2.3  & 59.8/55.1 \\
Online-Y & 10  & 0/63/122/2037 & 17.5/2.9  & 56.9/52.2 \\
Lan-Bridge & 11  & 0/63/122/2037 & 15.1/2.9  & 56.6/51.9 \\
Online-A & 5  & 0/63/122/2037 & 13.3/2.8  & 53.7/48.8 \\
NAIST-NICT-TIT & 13  & 22/63/122/2037 & 13.6/2.9  & 53.4/48.5 \\
AISP-SJTU & 13  & 0/63/122/2037 & 16.6/2.8  & 52.4/47.9 \\
KYB & 5  & 0/63/122/2037 & 13.3/2.7  & 31.8/26.6 \\
Online-G & 6  & 0/63/122/2037 & 12.4/2.8  & 25.0/20.2 \\
\bottomrule
\end{tabular}} \caption{en$\rightarrow$ja}
\end{subtable}

\caption{ Sentence Matching Statistics. ``Exact Match'' counts how many segments in the translation are identical to the corresponding segments of the reference translation. ``Self Mismatch'' provides various statistics: the number of non-unique source segments translated differently by the system/the number of non-unique source segments translated differently by the reference/the number of duplicated source segments/the total number of source segments. ``BLEU'' and ``COMET'' give the scores computed on the original/normalized translations.}
 \end{table}

    \begin{table}
\scriptsize 
\begin{subtable}[t]{0.49\textwidth}
\scalebox{0.85}{\begin{tabular}{rcccc}
\toprule
System & Exact Match & Self Mismatch & BLEU & COMET \\
\midrule
TartuNLP & 3  & 0/0/0/420 & 29.9/29.9  & -5.7/-5.7 \\
TAL-SJTU & 4  & 0/0/0/420 & 30.4/30.4  & -8.3/-8.3 \\
HuaweiTSC & 1  & 0/0/0/420 & 23.4/23.3  & -27.2/-27.2 \\
Liv4ever & 3  & 0/0/0/420 & 23.3/23.3  & -44.0/-44.0 \\
NiuTrans & 0  & 0/0/0/420 & 13.0/12.9  & -88.3/-88.3 \\
\bottomrule
\end{tabular}} \caption{liv$\rightarrow$en}
\end{subtable}
\begin{subtable}[t]{0.49\textwidth}
\scalebox{0.85}{\begin{tabular}{rcccc}
\toprule
System & Exact Match & Self Mismatch & BLEU & COMET \\
\midrule
TAL-SJTU & 0  & 0/0/0/420 & 17.0/16.9  & -29.5/-29.5 \\
TartuNLP & 1  & 0/0/0/420 & 15.0/14.3  & -36.8/-36.9 \\
HuaweiTSC & 1  & 0/0/0/420 & 12.8/12.9  & -38.8/-39.0 \\
Liv4ever & 0  & 0/0/0/420 & 14.7/14.5  & -39.4/-39.6 \\
NiuTrans & 0  & 0/0/0/420 & 12.3/11.4  & -81.9/-81.9 \\
\bottomrule
\end{tabular}} \caption{en$\rightarrow$liv}
\end{subtable}

\caption{ Sentence Matching Statistics. ``Exact Match'' counts how many segments in the translation are identical to the corresponding segments of the reference translation. ``Self Mismatch'' provides various statistics: the number of non-unique source segments translated differently by the system/the number of non-unique source segments translated differently by the reference/the number of duplicated source segments/the total number of source segments. ``BLEU'' and ``COMET'' give the scores computed on the original/normalized translations.}
 \end{table}

     \begin{table}
\scriptsize 
\begin{subtable}[t]{0.49\textwidth}
\scalebox{0.85}{\begin{tabular}{rcccc}
\toprule
System & Exact Match & Self Mismatch & BLEU & COMET \\
\midrule
Online-G & 132  & 0/0/5/2016 & 46.7/46.6  & 65.2/65.2 \\
JDExploreAcademy & 107  & 0/0/5/2016 & 45.1/45.1  & 64.9/64.9 \\
Online-Y & 98  & 0/0/5/2016 & 43.8/44.7  & 64.1/64.1 \\
Lan-Bridge & 98  & 0/0/5/2016 & 45.2/45.2  & 63.1/63.1 \\
Online-B & 97  & 0/0/5/2016 & 45.0/45.1  & 63.1/63.1 \\
Online-A & 96  & 0/0/5/2016 & 43.9/43.8  & 62.2/62.2 \\
Online-W & 71  & 2/0/5/2016 & 42.6/42.6  & 61.6/61.6 \\
HuaweiTSC & 97  & 0/0/5/2016 & 45.1/45.1  & 60.9/60.9 \\
SRPOL & 85  & 3/0/5/2016 & 43.6/43.6  & 59.5/59.5 \\
ALMAnaCH-Inria & 35  & 0/0/5/2016 & 30.3/30.3  & 26.9/26.9 \\
\bottomrule
\end{tabular}} \caption{ru$\rightarrow$en}
\end{subtable}
\begin{subtable}[t]{0.49\textwidth}
\scalebox{0.85}{\begin{tabular}{rcccc}
\toprule
System & Exact Match & Self Mismatch & BLEU & COMET \\
\midrule
Online-W & 81  & 27/25/122/2037 & 32.4/32.4  & 75.1/75.1 \\
Online-G & 74  & 2/25/122/2037 & 32.8/32.8  & 73.2/73.2 \\
Online-B & 100  & 0/25/122/2037 & 34.9/33.3  & 72.9/72.7 \\
Online-Y & 103  & 0/25/122/2037 & 33.2/32.1  & 69.8/69.6 \\
JDExploreAcademy & 81  & 1/25/122/2037 & 32.7/32.7  & 69.7/69.7 \\
Lan-Bridge & 83  & 0/25/122/2037 & 32.6/32.3  & 67.4/67.3 \\
Online-A & 70  & 0/25/122/2037 & 33.1/31.9  & 67.4/67.2 \\
PROMT & 73  & 0/25/122/2037 & 30.6/29.7  & 60.4/60.3 \\
SRPOL & 66  & 16/25/122/2037 & 30.4/30.4  & 59.8/59.8 \\
HuaweiTSC & 60  & 0/25/122/2037 & 30.8/30.8  & 59.2/59.2 \\
eTranslation & 58  & 0/25/122/2037 & 29.8/28.9  & 58.0/57.9 \\
\bottomrule
\end{tabular}} \caption{en$\rightarrow$ru}
\end{subtable}

\caption{ Sentence Matching Statistics. ``Exact Match'' counts how many segments in the translation are identical to the corresponding segments of the reference translation. ``Self Mismatch'' provides various statistics: the number of non-unique source segments translated differently by the system/the number of non-unique source segments translated differently by the reference/the number of duplicated source segments/the total number of source segments. ``BLEU'' and ``COMET'' give the scores computed on the original/normalized translations.}
 \end{table}

      \begin{table}
\scriptsize 
\begin{subtable}[t]{0.49\textwidth}
\scalebox{0.85}{\begin{tabular}{rcccc}
\toprule
System & Exact Match & Self Mismatch & BLEU & COMET \\
\midrule
Online-G & 7  & 0/0/14/1123 & 14.7/14.7  & -17.1/-17.2 \\
Lan-Bridge & 0  & 0/0/14/1123 & 15.3/14.0  & -48.0/-48.9 \\
\bottomrule
\end{tabular}} \caption{ru$\rightarrow$sah}
\end{subtable}
\begin{subtable}[t]{0.49\textwidth}
\scalebox{0.85}{\begin{tabular}{rcccc}
\toprule
System & Exact Match & Self Mismatch & BLEU & COMET \\
\midrule
Online-G & 32  & 0/1/15/1123 & 29.6/29.6  & 31.1/31.0 \\
Lan-Bridge & 0  & 0/1/15/1123 & 7.1/7.1  & -75.9/-75.7 \\
\bottomrule
\end{tabular}} \caption{sah$\rightarrow$ru}
\end{subtable}

\caption{ Sentence Matching Statistics. ``Exact Match'' counts how many segments in the translation are identical to the corresponding segments of the reference translation. ``Self Mismatch'' provides various statistics: the number of non-unique source segments translated differently by the system/the number of non-unique source segments translated differently by the reference/the number of duplicated source segments/the total number of source segments. ``BLEU'' and ``COMET'' give the scores computed on the original/normalized translations.}
 \end{table}

       \begin{table}
\scriptsize 
\begin{subtable}[t]{0.49\textwidth}
\scalebox{0.85}{\begin{tabular}{rcccc}
\toprule
System & Exact Match & Self Mismatch & BLEU & COMET \\
\midrule
Online-B & 101  & 0/0/0/2018 & 44.4/44.4  & 62.6/62.5 \\
Lan-Bridge & 106  & 0/0/0/2018 & 44.6/44.6  & 62.4/62.4 \\
Online-G & 73  & 0/0/0/2018 & 43.2/43.0  & 57.5/57.4 \\
Online-A & 76  & 0/0/0/2018 & 42.3/42.1  & 52.2/52.1 \\
HuaweiTSC & 71  & 0/0/0/2018 & 41.6/41.5  & 50.1/50.1 \\
Online-Y & 75  & 0/0/0/2018 & 41.8/41.0  & 49.8/49.7 \\
PROMT & 73  & 0/0/0/2018 & 42.1/42.1  & 49.6/49.6 \\
ARC-NKUA & 76  & 0/0/0/2018 & 41.9/41.9  & 49.6/49.6 \\
ALMAnaCH-Inria & 42  & 0/0/0/2018 & 30.0/29.7  & 21.8/21.7 \\
\bottomrule
\end{tabular}} \caption{uk$\rightarrow$en}
\end{subtable}
\begin{subtable}[t]{0.49\textwidth}
\scalebox{0.85}{\begin{tabular}{rcccc}
\toprule
System & Exact Match & Self Mismatch & BLEU & COMET \\
\midrule
Online-B & 75  & 0/6/122/2037 & 32.5/31.6  & 73.3/73.1 \\
Online-G & 36  & 2/6/122/2037 & 27.2/27.2  & 69.9/69.9 \\
Lan-Bridge & 71  & 0/6/122/2037 & 29.5/29.5  & 65.8/65.7 \\
Online-A & 53  & 0/6/122/2037 & 28.0/27.5  & 60.9/60.8 \\
eTranslation & 52  & 0/6/122/2037 & 26.2/25.2  & 54.6/54.4 \\
HuaweiTSC & 45  & 0/6/122/2037 & 26.5/26.3  & 54.4/54.4 \\
Online-Y & 49  & 0/6/122/2037 & 26.9/25.9  & 52.0/51.8 \\
ARC-NKUA & 50  & 0/6/122/2037 & 25.2/24.4  & 49.2/49.1 \\
\bottomrule
\end{tabular}} \caption{en$\rightarrow$uk}
\end{subtable}

\caption{ Sentence Matching Statistics. ``Exact Match'' counts how many segments in the translation are identical to the corresponding segments of the reference translation. ``Self Mismatch'' provides various statistics: the number of non-unique source segments translated differently by the system/the number of non-unique source segments translated differently by the reference/the number of duplicated source segments/the total number of source segments. ``BLEU'' and ``COMET'' give the scores computed on the original/normalized translations.}
 \end{table}

       \begin{table}
\scriptsize 
\begin{subtable}[t]{0.49\textwidth}
\scalebox{0.85}{\begin{tabular}{rcccc}
\toprule
System & Exact Match & Self Mismatch & BLEU & COMET \\
\midrule
Online-G & 15  & 0/1/4/1875 & 34.1/33.7  & 45.7/45.5 \\
JDExploreAcademy & 22  & 0/1/4/1875 & 37.9/37.9  & 45.2/45.2 \\
LanguageX & 22  & 0/1/4/1875 & 36.4/36.4  & 45.0/45.0 \\
Lan-Bridge & 21  & 0/1/4/1875 & 32.9/32.9  & 43.1/43.1 \\
HuaweiTSC & 24  & 0/1/4/1875 & 34.3/34.3  & 42.9/42.9 \\
Online-B & 18  & 0/1/4/1875 & 33.3/33.3  & 42.2/42.1 \\
AISP-SJTU & 15  & 0/1/4/1875 & 34.4/34.3  & 41.7/41.7 \\
Online-Y & 8  & 0/1/4/1875 & 31.1/30.8  & 40.9/40.7 \\
Online-A & 10  & 0/1/4/1875 & 31.6/31.7  & 35.3/35.2 \\
Online-W & 16  & 1/1/4/1875 & 28.1/28.1  & 31.7/31.7 \\
NiuTrans & 12  & 0/1/4/1875 & 30.5/30.5  & 31.4/31.3 \\
DLUT & 9  & 0/1/4/1875 & 29.0/29.0  & 30.7/30.7 \\
\bottomrule
\end{tabular}} \caption{zh$\rightarrow$en}
\end{subtable}
\begin{subtable}[t]{0.49\textwidth}
\scalebox{0.85}{\begin{tabular}{rcccc}
\toprule
System & Exact Match & Self Mismatch & BLEU & COMET \\
\midrule
LanguageX & 95  & 0/3/122/2037 & 24.9/4.9  & 63.9/60.6 \\
Online-B & 65  & 0/3/122/2037 & 48.1/9.2  & 61.8/58.5 \\
JDExploreAcademy & 59  & 1/3/122/2037 & 24.0/5.2  & 61.8/58.7 \\
Lan-Bridge & 49  & 20/3/122/2037 & 23.6/5.1  & 61.4/58.0 \\
Online-W & 32  & 36/3/122/2037 & 16.4/4.4  & 61.0/58.0 \\
Manifold & 47  & 0/3/122/2037 & 23.4/5.9  & 60.2/57.3 \\
Online-Y & 44  & 0/3/122/2037 & 28.0/6.0  & 59.8/56.5 \\
HuaweiTSC & 43  & 0/3/122/2037 & 25.3/4.9  & 59.6/56.5 \\
Online-A & 50  & 0/3/122/2037 & 26.4/6.3  & 57.4/53.8 \\
AISP-SJTU & 52  & 0/3/122/2037 & 25.7/4.8  & 56.6/53.2 \\
DLUT & 35  & 0/3/122/2037 & 21.1/5.2  & 52.2/48.5 \\
Online-G & 38  & 3/3/122/2037 & 16.1/5.0  & 51.3/48.3 \\
\bottomrule
\end{tabular}} \caption{en$\rightarrow$zh}
\end{subtable}

\caption{ \label{tab:lastmatch} Sentence Matching Statistics. ``Exact Match'' counts how many segments in the translation are identical to the corresponding segments of the reference translation. ``Self Mismatch'' provides various statistics: the number of non-unique source segments translated differently by the system/the number of non-unique source segments translated differently by the reference/the number of duplicated source segments/the total number of source segments. ``BLEU'' and ``COMET'' give the scores computed on the original/normalized translations.}
 \end{table}
\end{document}